\documentclass{bmvc2k}

\title{TriCCOT: Tri-part Convolutional Conformal Transformer for Onboard Space Object Detection}

\addauthor{Adrien Dorise}{adrien.dorise@cnes.fr}{1,2}
\addauthor{Marjorie Bellizzi}{marjorie.bellizzi@irt-saintexupery.com}{2}
\addauthor{Julia Cohen}{julia.cohen@cnes.fr}{1}
\addauthor{Stéphane May}{stephane.may@cnes.fr}{1}

\addinstitution{
 Centre National d'Etudes Spatiales, CNES \\
Toulouse, France}
\addinstitution{
 IRT Saint-Exupéry\\
 Toulouse, France
}

\runninghead{Dorise, Bellizzi, Cohen, May}{TriCCOT for onboard object detection}

\usepackage{booktabs}
\usepackage{graphicx}
\usepackage{subcaption}
\usepackage{float}
\usepackage{caption}
\usepackage{booktabs}
\usepackage{multirow}
\usepackage{adjustbox}
\usepackage{makecell,siunitx}
\usepackage{comment}
\usepackage{amssymb}
\usepackage{amsmath}
\usepackage{calc}
\usepackage{array}
\usepackage{wrapfig}

\begin{document}

\maketitle

\newcommand{\FigArchitecture}{%
\begin{figure}[htbp]
    \centerline{\includegraphics[width=\textwidth]{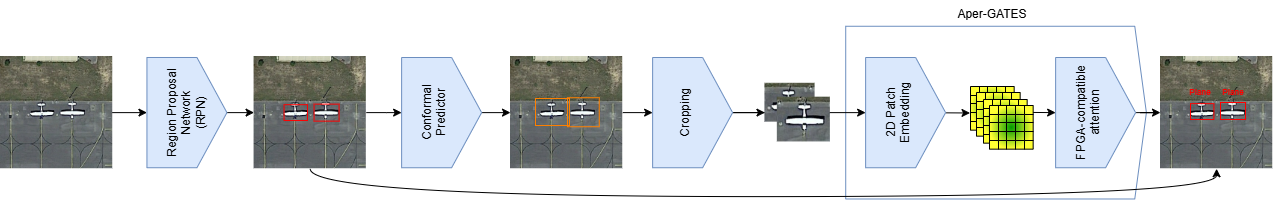}}
    \vspace{4pt}

    \caption{TriCCOT architecture. It is composed of a Region Proposal Network, a conformal predictor, and Aper-GATES, our convolutional self-attention classifier. The RPN predicts bounding boxes that are then refined using a conformal predictor. Cropped images are created from the conformal bounding boxes, which are aggregated into a 2D token space that serves as input for the attention-based classifier.}
    \label{fig:architecture}
\end{figure} 
}

\newcommand{\FigAperGates}{%
\begin{figure}[htbp]
    \centerline{\includegraphics[width=\textwidth]{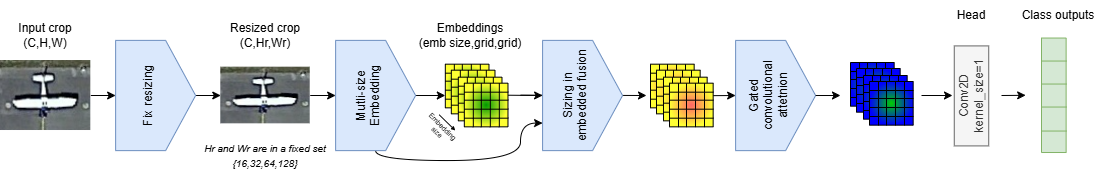}}
    \vspace{4pt}

    \caption{Aper-GATES architecture. It takes cropped images as input, maximising useful information. Cropped images are resized into pre-fixed sizes and projected into an embedding space. The attention matrix is computed using a gated convolutional attention module. Finally, a 2D convolutional mapping outputs the prediction.}
    \label{fig:apergates}
\end{figure} 
}

\newcommand{\FigEmbedding}{%
\begin{figure}[htbp]
    \centerline{\includegraphics[width=\textwidth]{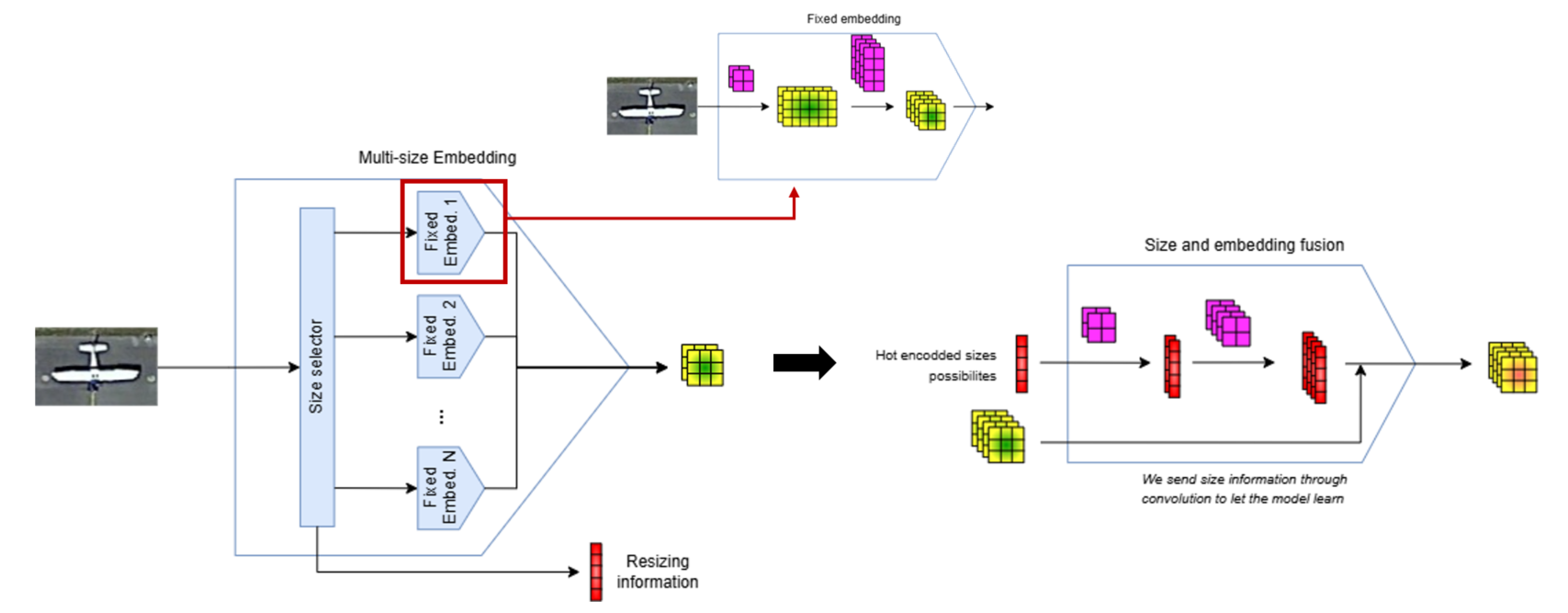}}
    \vspace{4pt}

    \caption{Multi-resolution patch embedding pipeline. The size selector is implemented using Boolean conditioning on the CPU. A fixed-size embedding space is achieved by carefully selecting the convolutional kernel for each admissible input resolution, which improves DPU deployment by keeping the data shape fixed throughout the model. Finally, the metadata are encoded into the embedding space.}
    \label{fig:embedded}
\end{figure} 
}

\newcommand{\FigGatedAttention}[1]{%
% {r} aligns it to the right, {l} to the left. 
% 0.5\textwidth is the width of the image area.
\begin{wrapfigure}{l}{0.5\textwidth}
    \centering
    \includegraphics[width=0.95\linewidth]{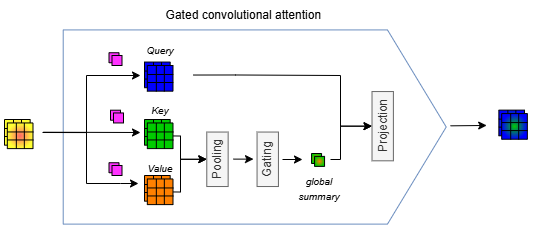}
    \caption{Aper-GATES attention mechanism. Context is gathered by a global descriptor.}
    \label{fig:gated_attention}
\end{wrapfigure}
#1
}

\newcommand{\FigDiorExample}{%
\begin{figure}[htbp]
    \centerline{\includegraphics[width=\textwidth]{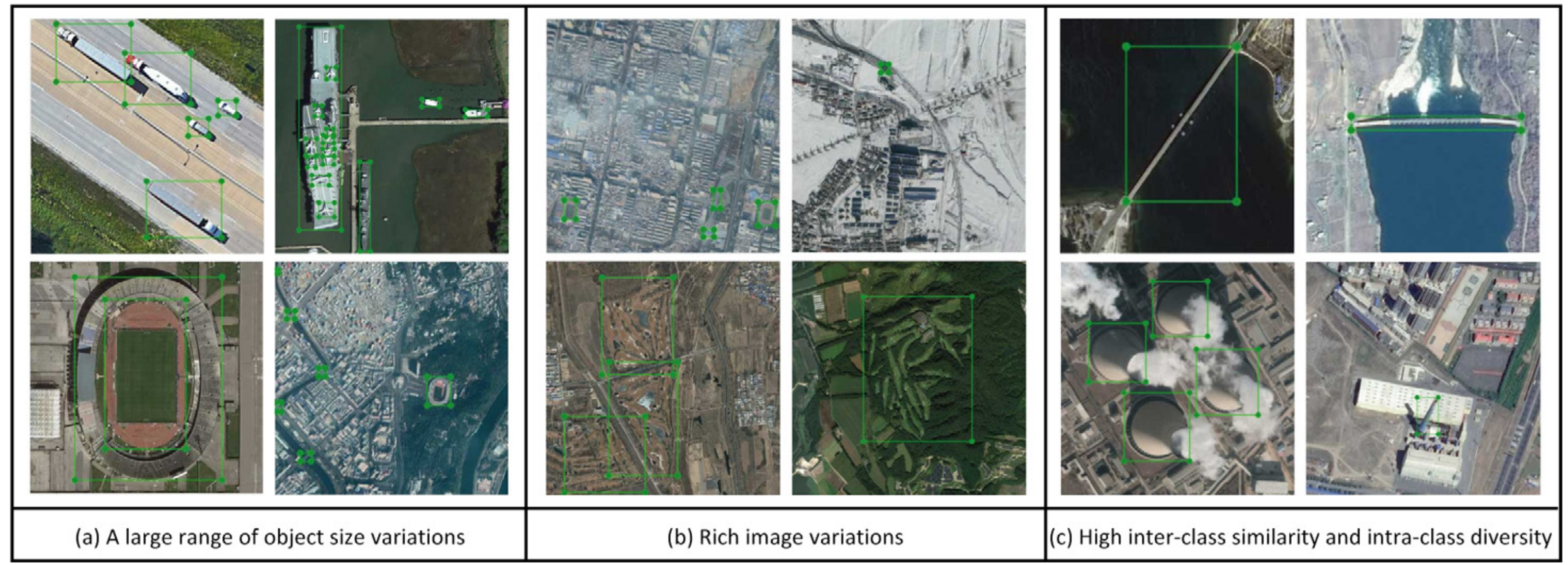}}
    \vspace{4pt}

    \caption{DIOR dataset characteristics \cite{DIOR}}
    \label{fig:dior_example}
\end{figure} 
}

\newcommand{\FigRawExample}{%
\begin{figure}[H]
\centering
\setlength{\tabcolsep}{2pt}   % small horizontal padding
\renewcommand{\arraystretch}{0.5}

% amount of vertical padding
\newcommand{\imgvpad}{0.1em}

% Outer table (border + separator)
\begin{tabular}{|c|c|}
\hline
% ================= LEFT GROUP (a) =================
\begin{subfigure}{0.47\textwidth}
    \centering
    \begin{tabular}{cc}
        \\
        \vspace{\imgvpad}
        \includegraphics[width=0.48\linewidth]{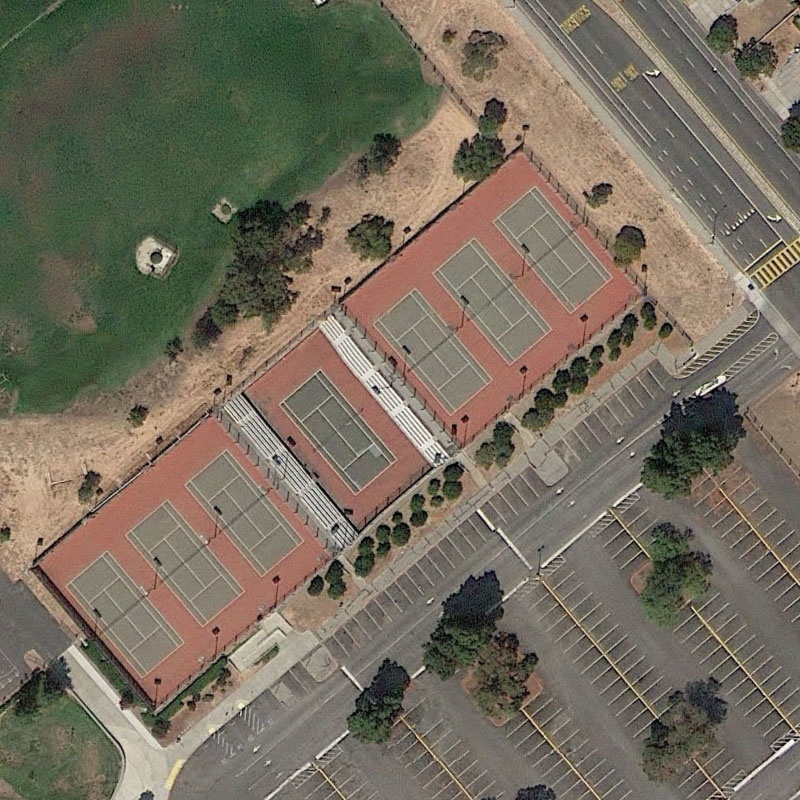}
        &
        \vspace{\imgvpad}
        \includegraphics[width=0.48\linewidth]{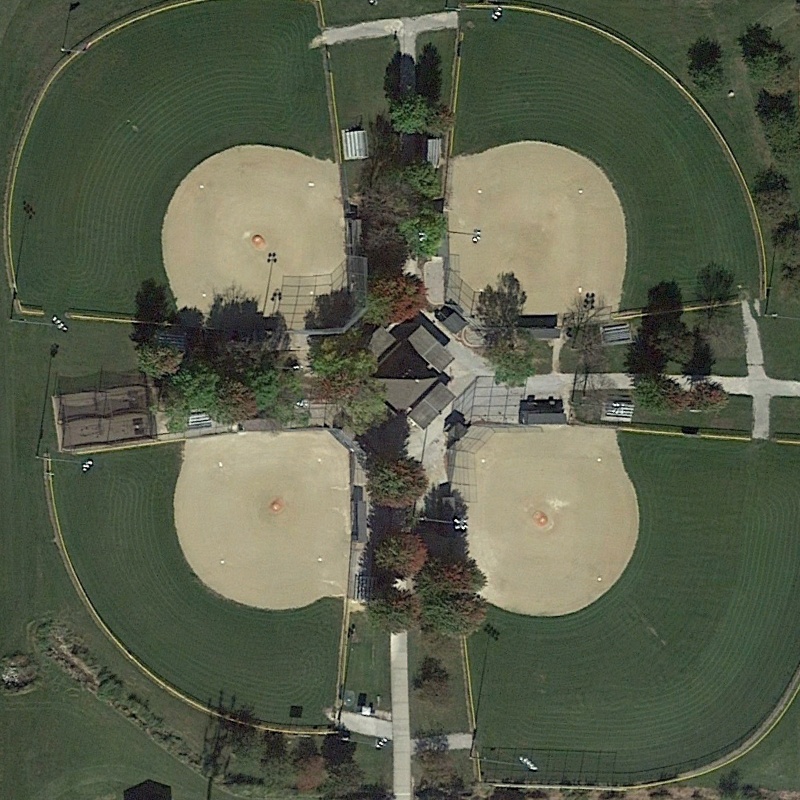}
        \\

        \vspace{\imgvpad}
        \includegraphics[width=0.48\linewidth]{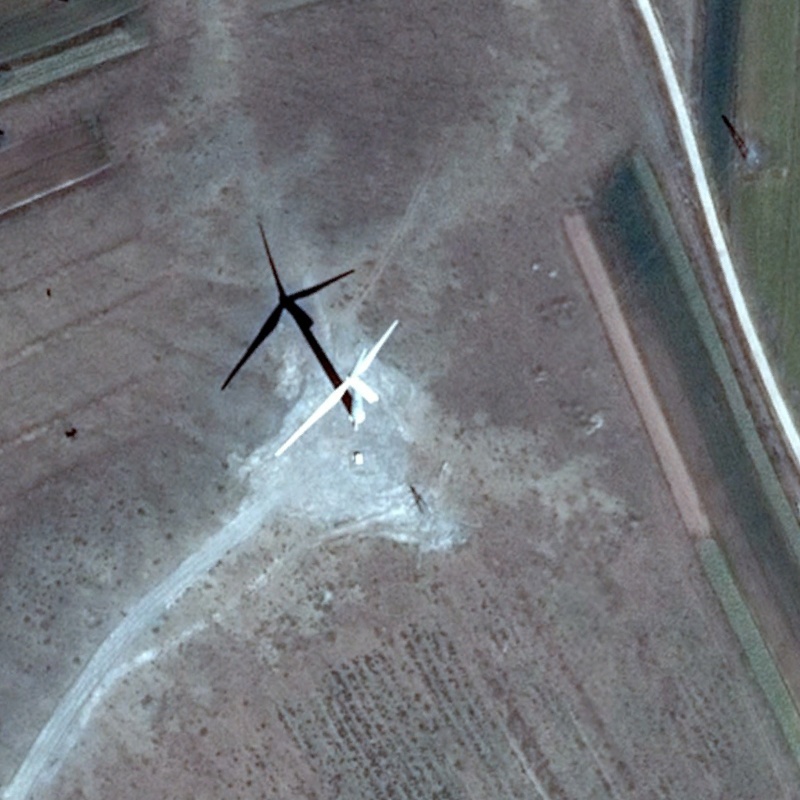}
        &
        \vspace{\imgvpad}
        \includegraphics[width=0.48\linewidth]{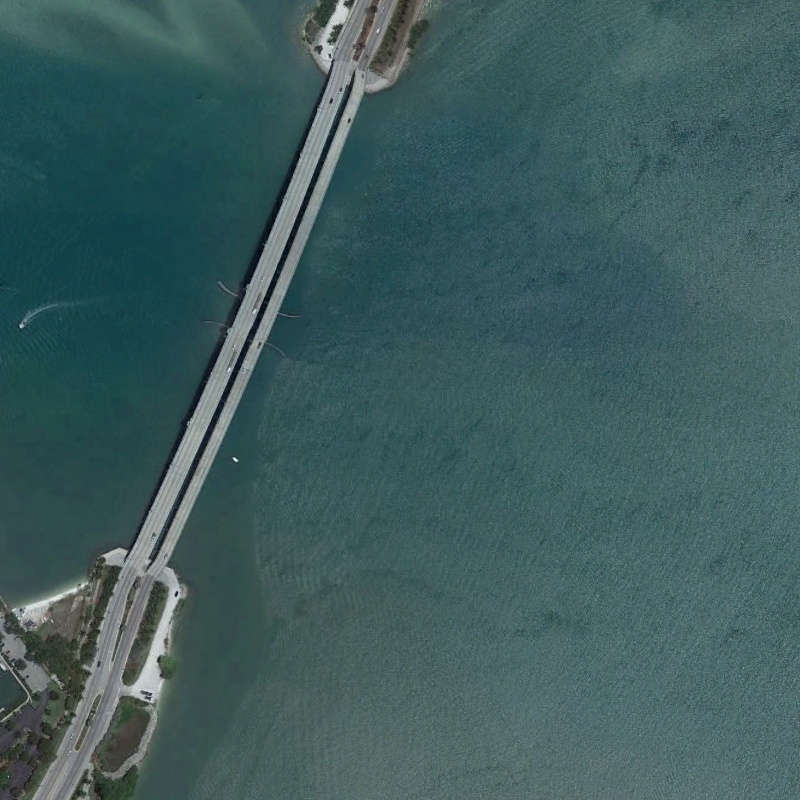}
        \vspace{\imgvpad}
        \\
        \hline
    \end{tabular}
    \caption{Original images}
\end{subfigure}

&
% ================= RIGHT GROUP (b) =================
\begin{subfigure}{0.47\textwidth}
    \centering
    \begin{tabular}{cc}
        \vspace{\imgvpad}
        \includegraphics[width=0.48\linewidth]{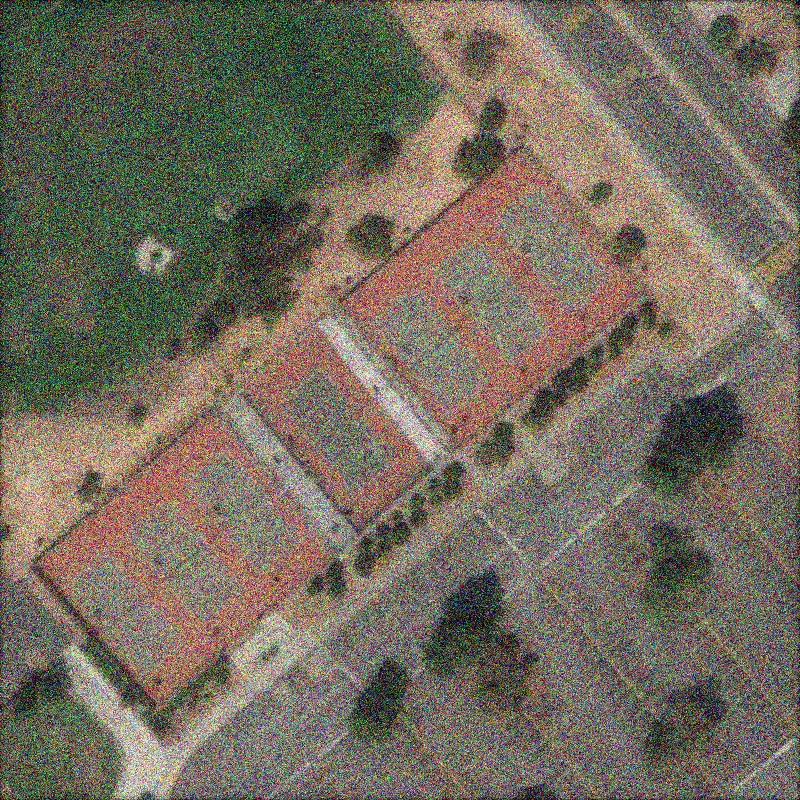}
        &
        \vspace{\imgvpad}
        \includegraphics[width=0.48\linewidth]{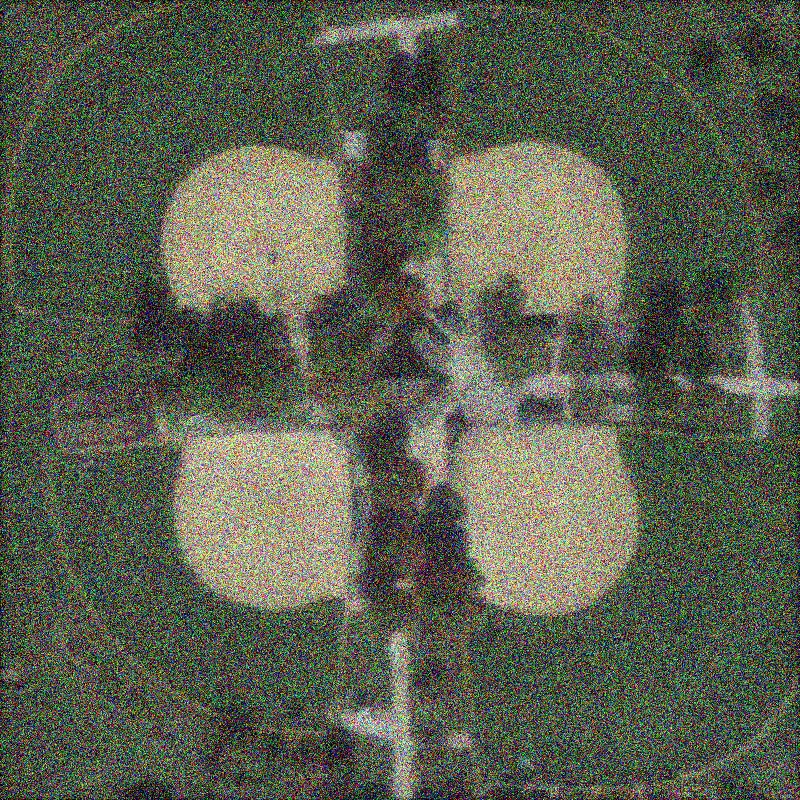}
        \\

        \vspace{\imgvpad}
        \includegraphics[width=0.48\linewidth]{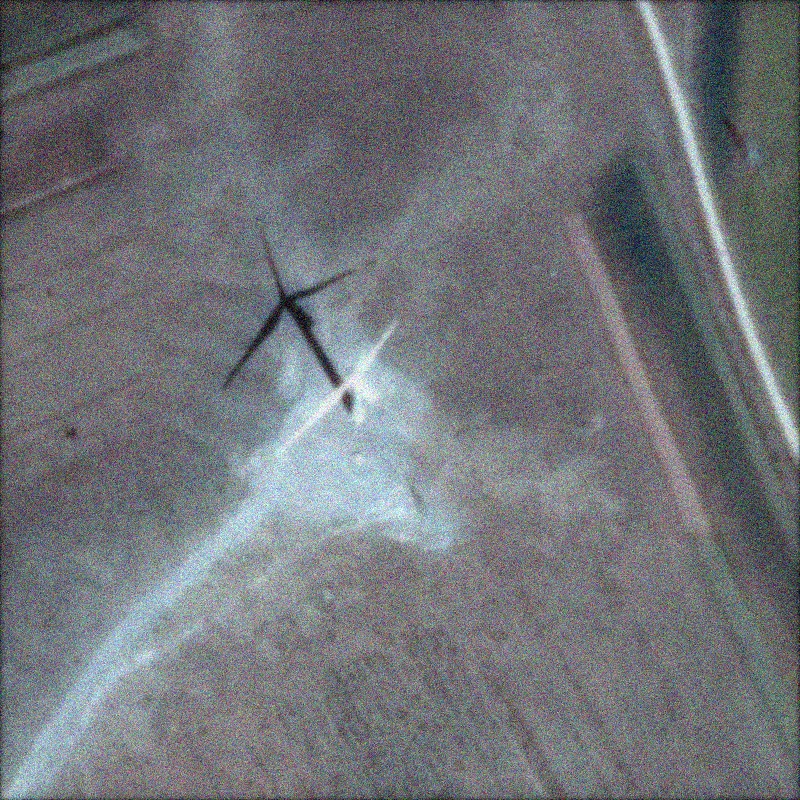}
        &
        \vspace{\imgvpad}
        \includegraphics[width=0.48\linewidth]{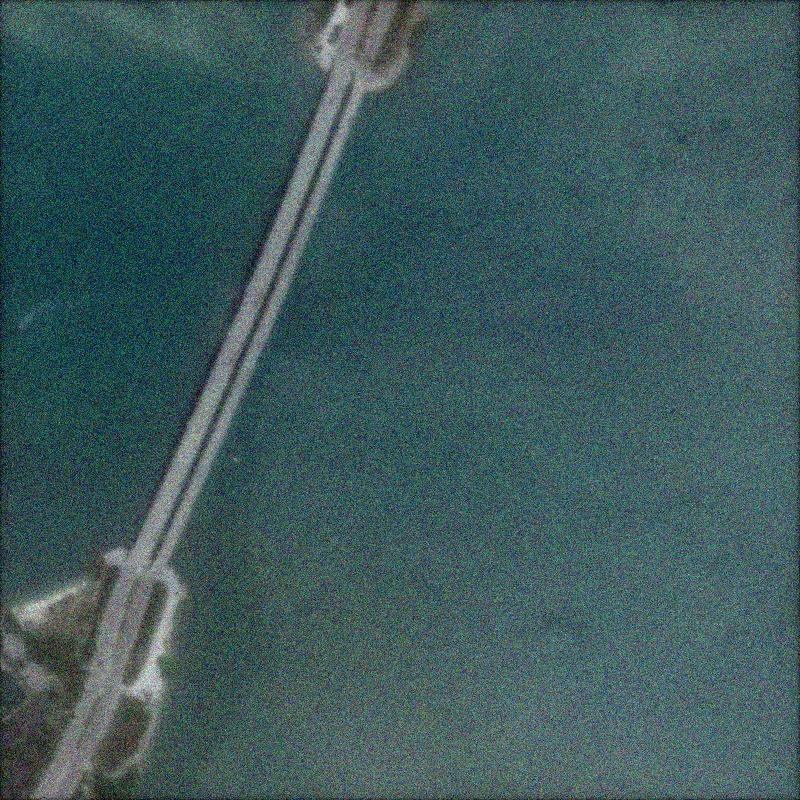}
        \vspace{\imgvpad}
        \\
        \hline
    \end{tabular}

    \caption{Noisy simulated images \cite{raw_detection_edhpc}}
\end{subfigure}

\\
\hline
\end{tabular}
\vspace{8pt}

\caption{Examples illustrating the DIOR dataset characteristics.}
\label{fig:raw_example}

\end{figure}
}

\newcommand{\FigPreds}{%
\begin{figure}[t]
    \centering
    \setlength{\tabcolsep}{2pt}
    \renewcommand{\arraystretch}{1.05}

    \begin{tabular}{
        |>{\centering\arraybackslash}m{0.30\linewidth}
        |>{\centering\arraybackslash}m{0.30\linewidth}
        |>{\centering\arraybackslash}m{0.28\linewidth}|
    }
    \hline
    \vspace{2pt}
    \textbf{Pred. on originals}
    &
    \vspace{2pt}
    \textbf{Pred. on degraded}
    &
    \vspace{2pt}
    \textbf{Conformal crops}
    \\[0.35em]
    \hline
    
    % Row 1
    \vspace{4pt}
    \includegraphics[width=0.86\linewidth]{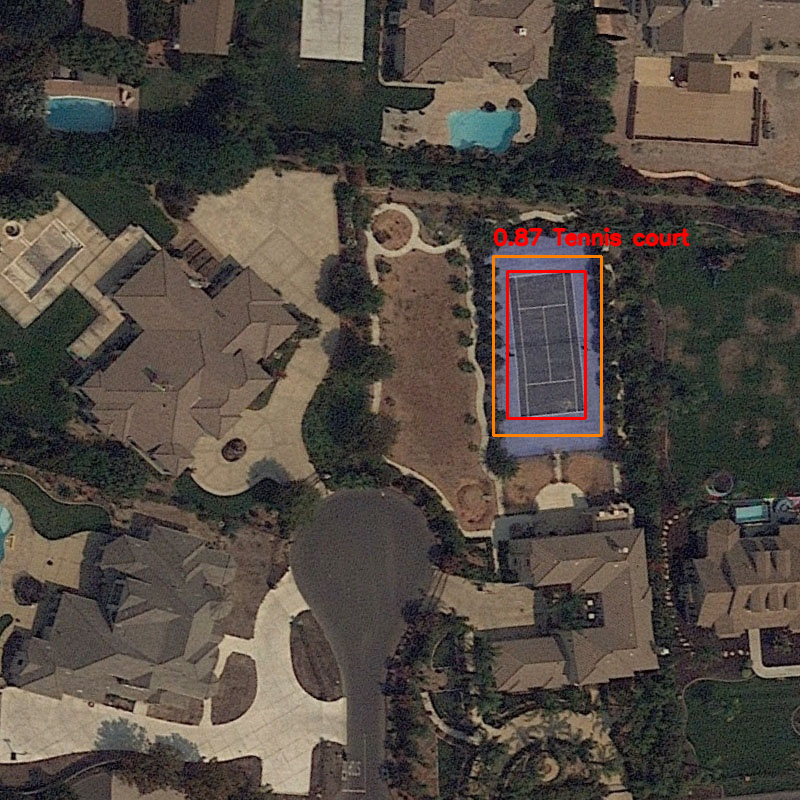}
    &
    \vspace{4pt}
    \includegraphics[width=0.86\linewidth]{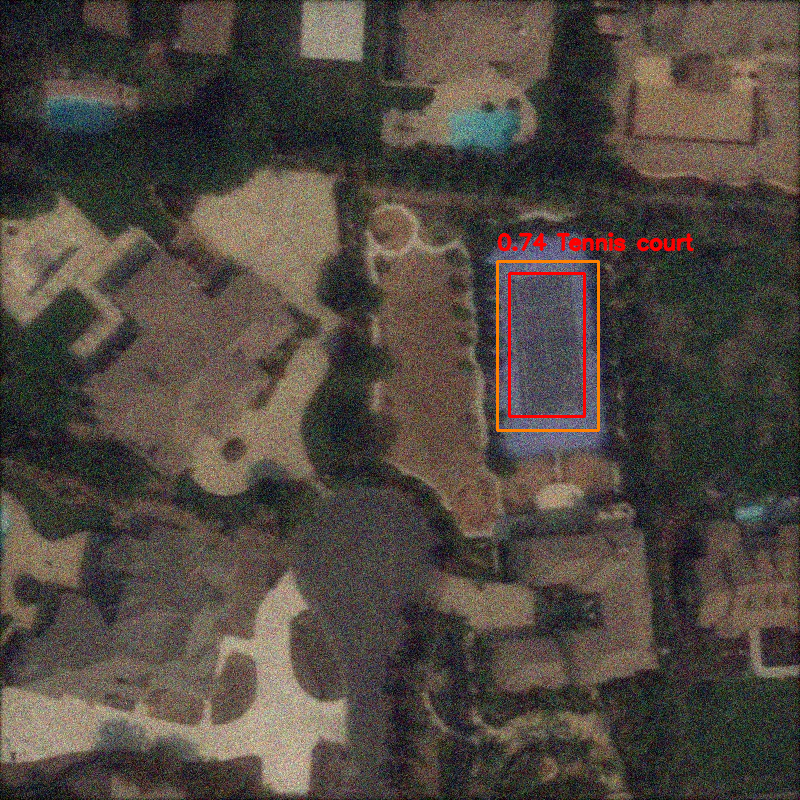}
    &
    \makebox[\linewidth][c]{%
        \includegraphics[width=0.39\linewidth]{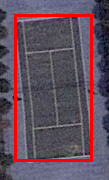}%
        \hspace{10pt}%
        \includegraphics[width=0.39\linewidth]{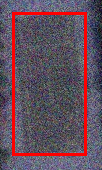}%
    }
    \\[0.45em]
    \hline
    % Row 2
    \vspace{4pt}
    \includegraphics[width=0.86\linewidth]{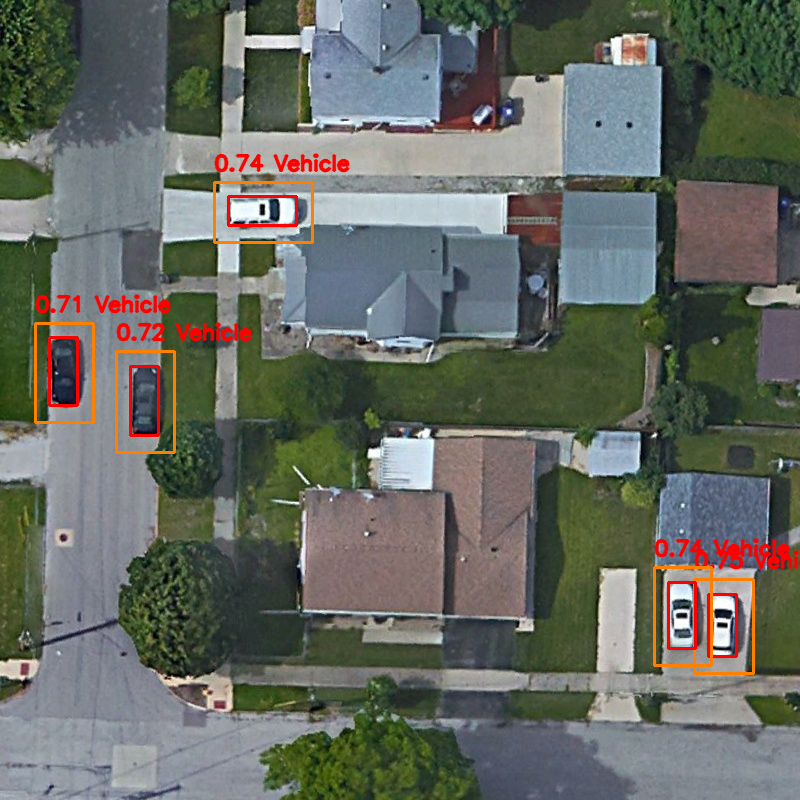}
    &
    \vspace{4pt}
    \includegraphics[width=0.86\linewidth]{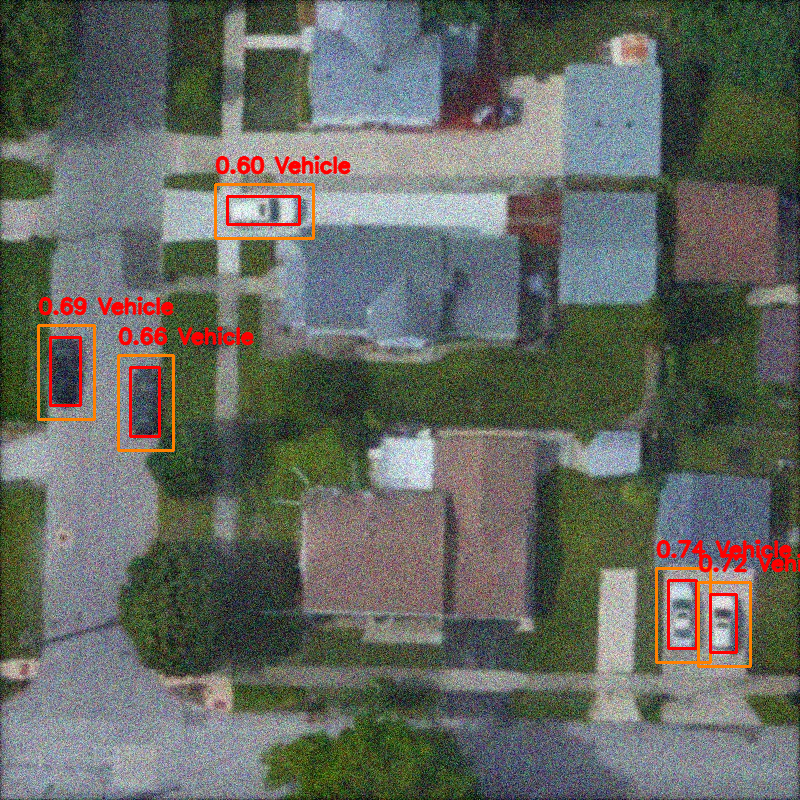}
    &
    \makebox[\linewidth][c]{%
    
        \includegraphics[width=0.39\linewidth]{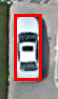}%
        \hspace{10pt}%
        \includegraphics[width=0.39\linewidth]{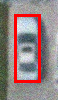}%
    }
    \\[0.45em]
    \hline

    % Row 3
    \vspace{4pt}
    \includegraphics[width=0.86\linewidth]{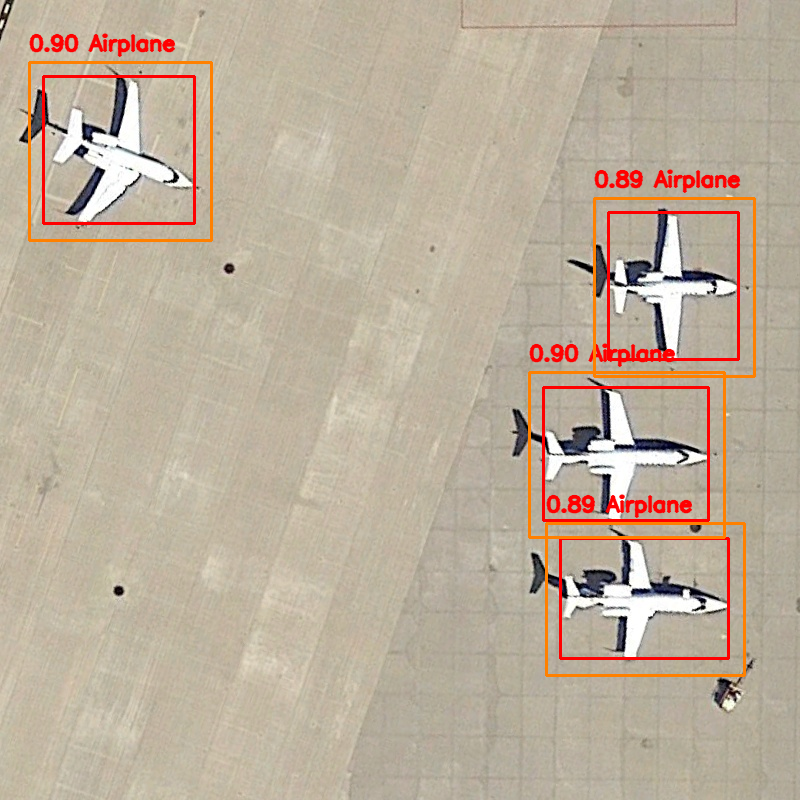}
    &
    \vspace{4pt}
    \includegraphics[width=0.86\linewidth]{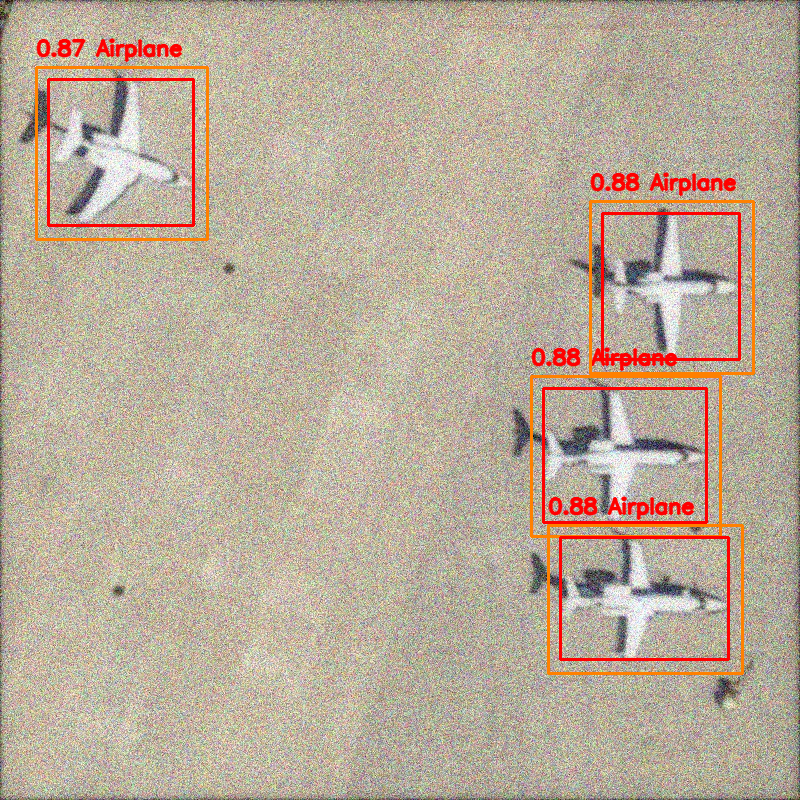}
    &
    \makebox[\linewidth][c]{%
        \includegraphics[width=0.39\linewidth]{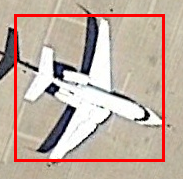}%
        \hspace{10pt}%
        \includegraphics[width=0.39\linewidth]{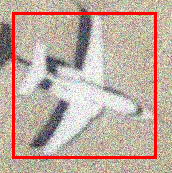}%
    }
    \\[0.45em]
    \hline

    \end{tabular}
\vspace{8pt}
    \caption{
    TriCCOT prediction on sampled original and degraded images. 
    }
    \label{fig:predictions_and_crops}
\end{figure}
}
\newcommand{\TabDiorBenchmark}{%
\begin{table}[t]
\centering
\caption{Benchmark on the original DIOR dataset.}
\vspace{8pt}
\label{tab:dior_benchmark}
\begin{adjustbox}{max width=\linewidth}
\begin{tabular}{lcccccc}
\toprule
\textbf{Method} & \textbf{mAP@50} & \textbf{Precision} & \textbf{Recall} 
& \textbf{FPS} $\uparrow$ & \textbf{Params (M)} & \textbf{FLOPs (G)} \\
\midrule
Faster R-CNN            & 78\% & 85\% & 75\% & 15  & 10 & 207 \\
YoloX-S                 & 80\% & 83\% & 77\% & 18  & 6 & 198 \\
Yolov11n                & 85\% & 88\% & 80\% & 20  & 3 & 190  \\
Yolov11l                & 78\% & 85\% & 75\% & 10  & 41.5 & 207  \\
DETR                    & 88\% & 90\% & 75\% & 4  & 50 & 250 \\
\textbf{TriCCOT (ours)} & 82\% & 85\% & 77\% & 18 & 3.5 & 190 \\
\bottomrule
\end{tabular}
\end{adjustbox}
\end{table}
}

\newcommand{\TabTriCCOTAblation}{%
\begin{table}[t]
\centering
\caption{Ablation study of the TriCCOT model on the DIOR dataset.}
\vspace{8pt}
\label{tab:triccot_ablation}
\begin{adjustbox}{max width=0.9\linewidth}
\begin{tabular}{lccccc}
\toprule
\textbf{Configuration} & \textbf{mAP@50} & \textbf{mAP@75} & \textbf{mAP@50:95} & \textbf{Precision} & \textbf{Recall} \\
\midrule
Full TriCCOT               & \textbf{76.49\%} &\textbf{ 40.08\%} & \textbf{45.51\%} & \textbf{76.21\%} & \textbf{53.79\%}  \\
w/o Aper-GATES           & 75.67\% & 39.94\% & 44.09\% & 70.19\% & 38.99\%  \\
w/o ConSmax                & 74.30\% & 39.09\% & 43.18\% & 68.97\% & 25.09\% \\
w/o Multi Res. Patch Embedding  & 72.55\% & 38.07\% & 41.55\% & 63.54\% & 20.76\% \\
w/o Conformal Pred.             & 69.10\% & 38.60\% & 42.67\% & 65.21\% & 25.51\% \\
\bottomrule
\end{tabular}
\end{adjustbox}
\end{table}
}

\newcommand{\TabBenchmark}{%
\begin{table*}[t]
\centering
\caption{Benchmark comparison on DIOR Original, DIOR Raw, and VDVRaw datasets.
Training times correspond to a single NVIDIA RTX 4090. Bold values indicate the best result among FPGA-compatible models, while underlined values indicate the best overall result.}
\label{tab:benchmark_all}
\vspace{6pt}

\begin{adjustbox}{max width=\textwidth}
\begin{tabular}{lccc|cc|cc|cc}
\toprule

& \multicolumn{3}{c|}{\textbf{Model Characteristics}}
& \multicolumn{2}{c|}{\textbf{DIOR Original} ($D_{l1}$)}
& \multicolumn{2}{c|}{\textbf{DIOR Raw} ($D_{\mathrm{raw}}$)}
& \multicolumn{2}{c}{\textbf{VDVRaw}} \\

\cmidrule(lr){2-4}
\cmidrule(lr){5-6}
\cmidrule(lr){7-8}
\cmidrule(lr){9-10}

\textbf{Method}
& \textbf{Params}
& \textbf{GFLOPs}
& \textbf{Train.}
& \textbf{mAP@50}
& \textbf{mAP@50:95}
& \textbf{mAP@50}
& \textbf{mAP@50:95}
& \textbf{mAP@50}
& \textbf{mAP@50:95} \\

\midrule
\multicolumn{10}{c}{\textit{Non-FPGA-compatible transformer models}} \\
\midrule

DETR \cite{DETR}
& 41.5\,M & 38.4 & $\sim$30\,h
& 49.9 & 28.7
& 49.1 & 27.7
& 8.6 & 5.6 \\

RT-DETR \cite{rt_detr}
& 20.0\,M & 60.0 & $\sim$30\,h
& \underline{82.1} & \underline{51.3}
& \underline{80.3} & \underline{48.2}
& \underline{29.7} & \underline{26.5} \\

\midrule
\multicolumn{10}{c}{\textit{FPGA-compatible models}} \\
\midrule

Faster R-CNN \cite{faster_RCNN}
& 40.0\,M & 89.0  & $\sim$8\,h
& 74.8 & 47.4
& 69.2 & 40.4
& 18.2 & 12.2 \\

YOLOX-S \cite{yolox}
& 8.9\,M & 26.8  & 4--5\,h
& \textbf{78.7} & 46.1
& 72.6 & 40.1
& 20.0 & 14.3 \\

NanoDet-Plus \cite{nanodet}
& 2.4\,M & 1.8  & 4--5\,h
& 75.3 & \textbf{47.9}
& 70.0 & 41.1
& 21.2 & 16.3 \\

\textbf{TriCCOT (ours)}
& 3.5\,M & 8.2  & 4--5\,h
& 76.5 & 45.5
& \textbf{74.8} & \textbf{42.2}
& \textbf{25.3} & \textbf{18.0} \\

\bottomrule
\end{tabular}
\end{adjustbox}
\end{table*}
}

\newcommand{\TabTrainingParams}{%
\begin{table}[ht]
\centering
\caption{Training and model parameters for the DIOR and VDVRaw datasets.}
\vspace{8pt}
\label{tab:parameters}
\begin{tabular}{llll}
\hline
\textbf{Param. Category} & \textbf{Parameter} & \textbf{DIOR} & \textbf{VDVRaw} \\
\hline

RPN 
& \textit{input size} & 640 & 640 \\
& \textit{max detections} & 70 & 70 \\

\hline

Conformal predictor
& \textit{$\alpha$} & 0.15 & 0.15 \\
& \textit{tested $\alpha$} & [0.10, 0.20] & - \\

\hline

Aper-GATES
& \textit{patch size} & \{16, 32, 64, 128\} & 32 \\
& \textit{embed. dim.} & 128 & 512 \\

\hline

Training
& \textit{epochs} & 200 & 200 \\
& \textit{lr} & $10^{-3}$ & $10^{-3}$ \\
& \textit{scheduler} & warmup. cosine & warm. cosine \\

\hline

Dataset
& \textit{nb. classes} & 20 & 17 \\
& \textit{train size} & 18\,000 & 3\,813 \\
& \textit{test size} & 2\,000 & 425 \\

\hline
\end{tabular}
\end{table}
}

\newcommand{\TabSimulationParameters}{%
\begin{table}[t]
  \centering
    \caption{Degradation parameters.}
    \vspace{8pt}

    \label{tab:simulation_parameters}

    \begin{adjustbox}{max width=\linewidth}
    \setlength{\tabcolsep}{10pt}  % default ≈ 6pt
    \begin{tabular}{lcc}
    \toprule
    \textbf{Set} &
    \makecell[c]{\textbf{MTF} \\ @Nyquist} &
    \makecell[c]{\textbf{(Lum., SNR) ref.}\\($W/m^{2}/sr/\si{\micro\metre} $, dB)} \\
    \midrule
    Set 1 & 0.05  & \makecell[c]{(25, 50);\\(100, 110)} \\
    Set 2 & 0.01  & \makecell[c]{(25, 15);\\(100, 30)} \\
    \bottomrule
  \end{tabular}
  \end{adjustbox}
\end{table}
}

\newcommand{\TabFPGABenchmarkFull}{%
\begin{table}[t]
\centering
\caption{Versal VCK190 inference performance on a $640\times640$ image. Latencies are averaged over 100 runs. Runner creation is performed only once at initialisation.}
\vspace{8pt}

\label{tab:fpga_benchmark_full}
\begin{adjustbox}{max width=\linewidth}
\begin{tabular}{lccccccc}
\toprule
\textbf{Method}
& \textbf{Hardware}
& \textbf{Prec.}
& \textbf{Size (MB)}
& \textbf{Latency (ms)} $\downarrow$
& \textbf{FPS} $\uparrow$
& \textbf{Power (W)} $\downarrow$
& \textbf{Pxls/s/W} $\uparrow$ \\
\midrule

Runner creation
& CPU
& -- & --
& 929.3 ($\pm$ 20.1)
& -- & -- & -- \\

Image preprocessing
& CPU
& -- & --
& 44.3 ($\pm$ 5.2)
& -- & -- & -- \\

\midrule
%\multicolumn{8}{l}{\textit{TriCCOT inference breakdown}} \\

\textbf{TriCCOT (ours)}
& CPU+DPU
& int8
& 10.0
& \textbf{35.5 ($\pm$ 3.4)}
& \textbf{28.2}
& 25
& \textbf{462\,824} \\

\quad \textit{RPN}
& \textit{DPU}
& \textit{int8} & \textit{1.9}
& \textit{29.1 ($\pm$ 3.0)}
& -- & -- & -- \\

\quad \textit{Conformal}
& \textit{CPU}
& \textit{float32} & \textit{2.9}
& \textit{2.4 ($\pm$ 0.2)}
& -- & -- & -- \\

\quad \textit{Size selector}
& \textit{CPU}
& --
& --
& \textit{0.1 ($\pm$ 0.0)}
& -- & -- & -- \\

\quad \textit{Aper-GATES}
& \textit{DPU}
& \textit{int8} & \textit{5.2}
& \textit{3.9 ($\pm$ 2.5)}
& -- & -- & -- \\

DETR \cite{DETR}
& CPU
& int8
& 162.7
& 13\,160 ($\pm$ 400)
& 0.08
& \textbf{22.5}
& 1\,456 \\

YOLOX-S \cite{yolox}
& DPU
& int8
& \textbf{9.2}
& 74.1 ($\pm$ 3.6)
& 13.5
& 25
& 221\,184 \\

\bottomrule
\end{tabular}
\end{adjustbox}
\end{table}
}

\newcommand{\TabPerClass}{
\begin{table}[t]
    \centering
    \caption{Per-class AP@50 of TriCCOT on the original DIOR dataset.}
    \vspace{8pt}
    
    \label{tab:per_class_dior}
    \small
    \setlength{\tabcolsep}{4pt}
    \renewcommand{\arraystretch}{0.9}
    \begin{tabular}{lc|lc}
        \hline
        \textbf{Class} & \textbf{AP@50} &
        \textbf{Class} & \textbf{AP@50} \\
        \hline
        Airplane                & 91.92 & Ground track field    & 83.43 \\
        Airport                 & 85.67 & Harbor                & 66.38 \\
        Baseball field          & 91.70 & Overpass              & 61.30 \\
        Basketball court        & 83.52 & Ship                  & 84.13 \\
        Bridge                  & 43.09 & Stadium               & 85.25 \\
        Chimney                 & 79.42 & Storage tank          & 73.07 \\
        Dam                     & 65.80 & Tennis court          & 91.67 \\
        Expressway service area & 92.42 & Train station         & 66.62 \\
        Expressway toll station & 73.67 & Vehicle               & 55.70 \\
        Golf field              & 79.57 & Windmill              & 84.34 \\
        \hline
    \end{tabular}
\end{table}

}

\newcommand{\TabPerClassSmall}{
\begin{table}[t]
    \centering
    \caption{AP@50 on representative DIOR categories.}
    \vspace{4pt}
    \label{tab:per_class_dior_small}
    \small
    \setlength{\tabcolsep}{3.5pt}
    \begin{tabular}{lcccccccc}
        \hline
        \textbf{Class} & Airplane & Baseball & Bridge & Harbor & Ship & Storage & Vehicle & Windmill \\
        \textbf{AP@50} & 91.92 & 91.70 & 43.09 & 66.38 & 84.13 & 73.07 & 55.70 & 84.34 \\
        \hline
    \end{tabular}
\end{table}
}

\begin{abstract}
Onboard object detection in Earth observation is constrained by limited computational resources and the absence of fully corrected imagery. While convolutional detectors are hardware-efficient, they often struggle to extract robust representations from raw and noisy data. Conversely, transformer-based models provide stronger global reasoning capabilities but remain difficult to deploy on FPGA accelerators due to quadratic attention complexity and non-compatible operations.

We introduce TriCCOT, a tri-part architecture for robust and deployable onboard object detection. TriCCOT combines a convolutional region proposal network, a conformal prediction stage, and Aper-GATES, our hardware-friendly attention-based classifier. The region proposal network generates candidate bounding boxes, which are subsequently enlarged via conformal prediction, providing a distribution-free probabilistic coverage guarantee. The resulting crops are processed by Aper-GATES, which reformulates self-attention through convolutional projections, global channel statistics, and hardware-friendly gating operations, avoiding standard transformer operations that are poorly suited to CNN-oriented accelerators.

Experiments on the DIOR and VDVRaw datasets demonstrate competitive detection performance and improved robustness to spatial blur and signal-dependent noise when compared to FPGA-compatible architectures. Finally, we report full deployment on a Xilinx Versal VCK190 FPGA without modifying the underlying DPU architecture, enabling unified CNN-Transformer inference for spaceborne embedded applications.

\end{abstract}
   
\section{Introduction}
\label{sec:intro}

Earth observation systems rely extensively on image processing techniques to transform raw, noisy acquisitions into exploitable products delivered to end users. Radiometric and geometric corrections are traditionally performed on the ground after data transmission from the satellite. Although effective, these processing chains are computationally intensive and time-consuming. At the same time, the steady increase of onboard computational resources has encouraged a paradigm shift toward embedding decision-making capabilities directly within the satellite platform~\cite{remote_sensing_review}. Tasks such as cloud detection and masking~\cite{cloud_detection_onboard}, image classification~\cite{opssat_meoni}, anomaly detection~\cite{irma_imagini}, and object detection~\cite{CIAR, raw_detection_edhpc} are increasingly being deployed onboard to reduce downlink bandwidth and latency.

Onboard models do not benefit from the fully corrected imagery available in ground-based processing pipelines. A common strategy is to reproduce part of the restoration workflow onboard, either through simplified physical corrections or lightweight neural approximations~\cite{pyraws_thraws, AI4SPACE_CONVBEERS}. However, such approaches introduce additional computational overhead and partially defeat the purpose of moving intelligence to the edge. An alternative direction is to directly exploit raw satellite imagery as model input, thereby eliminating preprocessing stages. This strategy remains marginal, because raw data are scarcely available \cite{pyraws_thraws}, and performance degradation is commonly observed when learning from uncorrected data~\cite{pyraws_maritime1,pyraws_maritime2}. In particular, previous studies have shown a noticeable decline in object detection accuracy when operating on raw images instead of processed ones~\cite{raw_detection_edhpc, AI4SPACE_CONVBEERS}, as convolutional architectures struggle to capture inter-class variations under degraded conditions.

Object detection nevertheless remains a central task in remote sensing applications~\cite{vessel_detection_yolo11_sentinel2}. Convolutional detectors have long dominated the field, while more recent transformer-based architectures have demonstrated superior representational capabilities in computer vision. Yet, attention mechanisms introduce significant computational challenges for onboard deployment. Softmax operations, high-dimensional matrix multiplications, and quadratic complexity with respect to the token count make them difficult to integrate into constrained hardware environments~\cite{attention_all_you_need}. 

In light of these considerations, we argue that the attention mechanism is efficient at finding discriminative representations from raw satellite imagery, provided it can be reformulated to meet strict deployment constraints. In this work, we introduce \textbf{Aper-GATES} (Aperture-Gated Attention Transformer for Embedded Systems), an efficient attention formulation compatible with FPGAs accelerators. Taking advantage of this new model, we also introduce \textit{TriCCOT} (Tri-Part Convolutional Conformal Transformer). This novel object detection architecture combines convolutional priors, conformal prediction principles, and Aper-GATES. The proposed approach is designed to operate directly on raw imagery while preserving computational feasibility, thereby bridging the gap between detection performance and embedded deployment requirements.

\section{Related Work}

\paragraph{Object detection in remote sensing}
    Object detection in remote sensing has been extensively investigated over the past decade. Conventional detection frameworks include both convolutional two-stage architectures \cite{RCNN, faster_RCNN, SPPNet} and one-stage variants~\cite{FCOS, yolo, yolox}, often reinforced with multi-scale feature pyramids to better capture objects of varying sizes~\cite{FPN_feature_pyramid_network}. Transformer-based detectors such as DETR~\cite{DETR} make use of the attention mechanism \cite{attention_all_you_need} and achieve strong performance, albeit at a higher computational cost. While these methods focus on detection accuracy under standard imaging conditions, they generally assume access to preprocessed data and do not explicitly address the constraints imposed by degraded or raw onboard imagery.

\paragraph{Onboard deployments for remote sensing applications}
The deployment of deep learning models directly onboard satellites has recently gained momentum as a means of reducing transmission overhead and enabling real-time responses~\cite{remote_sensing_review}. Embedded inference has been demonstrated for cloud detection~\cite{cloud_detection_onboard}, anomaly detection~\cite{onboard_oil_spill_detection}, and scene classification~\cite{opssat_meoni}. Several studies have explored learning from raw or minimally restored images to bypass computationally demanding correction pipelines~\cite{pyraws_maritime2,pyraws_thraws}. However, studies report performance drops when models are trained and evaluated on degraded data~\cite{raw_detection_edhpc}, highlighting the difficulty of extracting robust representations without full radiometric and geometric processing. Rather than improving preprocessing stages, our approach tackles this limitation at the architectural level.
%Detecting small objects remains particularly challenging in aerial imagery. Classical IoU-based regression losses~\cite{IoU} tend to penalise localisation errors disproportionately for small bounding boxes, which has motivated the development of scale-aware formulations such as SIoU~\cite{SIoU}, which proved to be efficient for remote sensing applications \cite{SIoU_in_aerial_image,pyraws_maritime2}.

\paragraph{Embedded object detection}
Although transformer-based vision models offer strong representation learning \cite{DETR}, their attention mechanisms remain difficult to deploy on resource-constrained onboard accelerators. Hybrid architectures such as CvT~\cite{CvT_convolutional_vision_transformer} and MT~\cite{CMT_convolutional_NN_meets_transformers} incorporate convolutional inductive biases within attention-based models. Nevertheless, standard self-attention mechanisms~\cite{attention_all_you_need} scale quadratically with token number and rely on softmax normalisation, while linear approximations~\cite{linear_attention} still require global matrix operations. Such characteristics limit their compatibility with resource-constrained onboard hardware. Existing FPGA deployments in remote sensing have therefore largely concentrated on convolutional networks using platforms such as Xilinx Versal and DPU architectures~\cite{versal_dpu_for_CNN,versal_ship_detection_yolo,versal_plane_detection_yolo,versal_spaice_project}. In contrast, transformer acceleration typically demands specialised datapaths or dedicated quantisation-aware schemes~\cite{autoVit_fpga_accelerator,fpga_vit_accelerator_by_mapping_datapath, fpga_transformer_accelerator}. By reformulating attention without explicit token-token affinity computation and expressing it through convolutional projections combined with global channel statistics, our method enables integration within a standard CNN-oriented acceleration pipeline, reconciling transformer expressiveness with strict hardware constraints.

\paragraph{Conformal prediction}
Conformal prediction has recently emerged as a principled tool for providing distribution-free uncertainty guarantees~\cite{conformal_gentle_introduction}. Extensions to object detection enable the construction of bounding boxes with probabilistic coverage guarantees~\cite{conformal_andeol,conformal_paper}, primarily focusing on calibration and reliability. In contrast, we leverage conformal prediction as a structural component of the detection pipeline, systematically enlarging region proposals before classification. This design ensures contextual completeness and mitigates biases introduced during proposal generation, without resorting to heuristic scaling rules.

\section{Combining convolution, conformal and attention}
\label{sec:materials}

\subsection{Model overview}
\label{sec:model}
TriCCOT comprises three sequential stages: a convolutional Region Proposal Network (RPN), conformal adjustment of the predicted boxes, and crop classification with Aper-GATES, our FPGA-friendly attention model. Processing object-centric crops reduces irrelevant image content and enables a compact classifier. The complete architecture is shown in Fig.~\ref{fig:architecture}.

\FigArchitecture

    \subsection{Region Proposal Network (RPN)}
    \label{sec:RPN}
    The core concept of the region proposal network derives from two-stage detection models \cite{RCNN} that rely on a first detector to predict bounding boxes and a second classifier to predict the class of the detected objects. Our RPN is based on a CSPDarknet backbone coupled with an FPN mechanism \cite{FPN_feature_pyramid_network} to adapt to multiscale objects.
    
    As studies have shown that degradations introduced into raw data have little effect on bounding box localisation \cite{raw_detection_edhpc}, few modifications are required to the RPN, and other convolutional backbones can be used if needed.

    \subsection{Conformal predictor}
    \label{sec:conformal}

    RPN bounding boxes can be imperfect, leading to cropped or shifted objects that lack crucial information for classification. In addition, ground-truth annotations are typically tight around the object and may therefore provide little contextual information when directly converted into image crops. We therefore introduce a conformal predictor to relax the box constraints before feeding into the transformer.

    When applied to object detection, a conformal predictor aims to predict bounding boxes with probabilistic guarantees \cite{conformal_paper, conformal_andeol}. For a bounding box $Y=(Y^{x_{\min}},Y^{y_{\min}},Y^{x_{\max}},Y^{y_{\max}})$, the calibrated conformal predictor adjusts each predicted coordinate independently as
    \begin{equation}
    \label{eq:conformal}
        \widehat{C}_\alpha
        =
        \left\{
        \text{box with coordinates }
        \widehat{Y}^{\,j}_{\text{new}} - d^{\,j}_{\alpha/4}
        \ \text{for } 
        j \in \{x_{\min}, y_{\min}, x_{\max}, y_{\max}\}
        \right\}
    \end{equation}
    where $d^{\,j}_{\alpha/4}$ is the calibrated correction associated with coordinate $j$. These corrections are estimated independently for each boundary and may therefore differ in both magnitude and sign, allowing the conformal predictor to compensate not only for insufficient box size but also for systematic localisation biases.

    Applying conformal prediction to the RPN outputs ensures that, under the conformal coverage assumptions, a fraction $1-\alpha$ of ground-truth objects are entirely contained within the adjusted predicted bounding boxes. The resulting crops consequently provide additional contextual information to the attention-based classifier. Compared with a fixed multiplicative expansion ratio, conformal prediction provides a data-driven adjustment that accounts for the localisation errors and directional biases of the RPN with limited computational overhead.

    In this work, conformal prediction is applied by using the PUNCC library \cite{puncc_lib, puncc_paper}.

    \subsection{Aper-GATES: An Aperture-Gated Attention Transformer for Embedded Systems}
    \label{sec:vit}
        Transformer architectures have demonstrated strong performance across a wide range of tasks, primarily due to their ability to model long-range dependencies. Unlike convolutional neural networks, which are limited to the receptive fields of their kernels, Vision Transformers (ViTs) can aggregate features globally across an image. We believe that this global context can improve model robustness on degraded images.

        \FigAperGates
        
        However, deploying transformers on edge hardware (particularly FPGAs) remains non-trivial due to the high cost of the softmax operation and the quadratic complexity of the attention mechanism. To address these constraints, we propose Aper-GATES (Aperture-Gated Attention Transformer for Embedded Systems). Inspired by Squeeze-and-Excitation and gated approaches \cite{ConvVit, CvT_convolutional_vision_transformer, gated_network}, Aper-GATES utilises a hardware-efficient gating mechanism (analogous to an optical aperture) to modulate the flow of global information. Unlike SE-Net and GC-Net, which pool the feature map itself and reinject the resulting descriptor, Aper-GATES pools key-value interactions to obtain a global correlation descriptor that multiplicatively gates the query tensor $Q$. Aper-GATES trades pairwise token expressivity for architectural simplicity, numerical stability, and compatibility with FPGA deployment,  while preserving global context aggregation. The global architecture of Aper-GATES is displayed in Fig.~\ref{fig:apergates}. First, the crop goes through a dynamic resizing step. The resulting image is embedded in a latent space and processed by a gated convolutional attention mechanism. Finally, the attention result is processed by a convolutional layer that outputs the class probabilities. 

        \subsubsection{Multi-resolution patch embedding bank with fixed grid}
            Transformers can adapt to various input sizes. In the field of Large Language Models (LLMs), self-attention operates over variable-length token sequences in a latent space \cite{attention_all_you_need, causal_attention_in_VLM}. Conversely, ViTs are traditionally trained with a fixed input resolution (usually 224x224) \cite{DETR, ViT_vs_CNN_biases, MAE_masked_autoencoders}. Nevertheless, some studies have leveraged this property to support variable spatial scales, such as FlexiViT, Swin Transformer, or DINOv2 \cite{flexivit, swin_transformer, dinov2}. In practice, varying the image resolution generally changes the number of input tokens or requires positional-embedding interpolation, which is undesirable for static FPGA deployment. Since Aper-GATES takes cropped images of variable sizes from the conformal predictor as input, it is important to retain the size and shape information of the cropped objects. However, supporting variable input sizes can degrade FPGA performance. Therefore, we introduce a multi-resolution patch embedding that resizes the cropped images into fixed shapes, as displayed in Fig.~\ref{fig:embedded}.

            \FigEmbedding
            
            Let the input crop be denoted by $\mathbf{x} \in \mathbb{R}^{B \times 3 \times H \times W}$, where the spatial size $(H,W)$ is selected from a predefined set of admissible resolutions and is closest to the original crop size while preserving aspect ratio. In our implementation $H,W \in \{16,32,64,128\}$. Rather than using a single patch embedding layer, we define a bank of convolutional patch embedders, each specialised to a single admissible input size. For an input of size \((H,W)\), the corresponding branch is selected. Because this selection relies on Boolean conditions, it is outsourced to the CPU rather than executed on the FPGA DPU. Since this operation remains fast, it does not significantly degrade inference speed.  Each branch maps the input crop to a fixed spatial grid of size $G \times G$, with \(G=8\) in our implementation, allowing a single model to be used for all admissible input sizes. This is achieved via two strided convolutional stages: 
            \begin{equation}
            \mathbf{z}
            =
            \mathrm{Embed}_{H,W}(\mathbf{x})
            \in
            \mathbb{R}^{B \times D \times G \times G}.
            \end{equation}
            where the total downsampling stride is
            \begin{equation}
            s_h = \frac{H}{G}, \quad s_w = \frac{W}{G}
            \end{equation}
            For example, strides \(2,4,8,16\) are split as \((1,2)\), \((2,2)\), \((2,4)\), and \((4,4)\), respectively. This allows all supported input sizes to be projected to the same \(8\times8\) feature grid.

            To provide the encoder with explicit knowledge of the input geometry, we compute a normalised metadata vector $\mathbf{m} = [\mathbf{g}_{\mathrm{branch}}, \mathbf{m}_{\mathrm{size}}]$, where \(\mathbf{g}_{\mathrm{branch}}\) is the one-hot encoded branch selection vector and $\mathbf{m}_{\mathrm{size}}$ is a normalised size vector computed as:
            \begin{equation}
            \mathbf{m}_{\mathrm{size}}
            =
            \left[
            \frac{H}{H_{\max}},
            \frac{W}{W_{\max}},
            \frac{H/G}{P_{h,\max}},
            \frac{W/G}{P_{w,\max}}
            \right].
            \end{equation}
            where $P_{h,\max}$ and $P_{w,\max}$ denote the maximum patch-grid height and width allowed by the size selector, respectively. This metadata is projected via a two-layer convolutional network and integrated into the feature map $\mathbf{z}$ via additive residual fusion: $\mathbf{z}' = \mathbf{z} + \mathrm{CONV}(\mathbf{m})$. This ensures that, while the feature grid is fixed, the network remains aware of the original object scale.

            \subsubsection{Gated Convolutional Attention}
            Let $\mathbf{X} \in \mathbb{R}^{B \times C \times H \times W}$ be an intermediate feature map, where $B$ denotes the batch size, $C$ the number of channels, and $H \times W$ the spatial resolution. We use the multihead strategy to obtain DPU-compatible batch sizes by partitioning the channels into $h$ heads of dimension $d$, where $C = h d$. 
            
            \FigGatedAttention
             
            While convolutional projections for $\mathbf{Q, K, V}$ have been explored in prior works such as CvT \cite{CvT_convolutional_vision_transformer} and CMT \cite{CMT_convolutional_NN_meets_transformers} to introduce spatial inductive bias, these architectures typically construct token–token affinity matrices that retain quadratic attention complexity $\mathcal{O}(N^2)$. In contrast, the core of Aper-GATES is a Gated Convolutional Attention that trades pairwise interactions for a global second-order statistic, as displayed in Fig.~\ref{fig:gated_attention}, significantly reducing the computational footprint.

            \paragraph{Convolutional Projections}
            Instead of flattening spatial dimensions into $N = HW$ tokens, we preserve the two-dimensional structure and compute head-wise projections using grouped $1\times1$ convolutions:
            \begin{equation}
            \mathbf{Q} = \mathrm{Conv}_{1\times1}^{(g=h)}(\mathbf{X}), \quad
            \mathbf{K} = \mathrm{Conv}_{1\times1}^{(g=h)}(\mathbf{X}), \quad
            \mathbf{V} = \mathrm{Conv}_{1\times1}^{(g=h)}(\mathbf{X}),
            \end{equation}
            with $\mathbf{Q}, \mathbf{K}, \mathbf{V} \in \mathbb{R}^{B \times C \times H \times W}$.
            This operation maintains the spatial grid while projecting the input into the query, key, and value manifolds within each head's subspace.
            
            \paragraph{Global Second-Order Context}
            Rather than calculating an $N \times N$ affinity matrix, we take inspiration from the  Squeeze-and-Excitation (SE) \cite{gate_conv_network,gated_network} and Global Context (GC) \cite{non_local_nn} Networks, and ``squeeze'' the spatial interaction into a global channel descriptor $\mathbf{G} \in \mathbb{R}^{B \times C \times 1 \times 1}$ via global average pooling ($\mathcal{P}$) of the Key-Value product:
            \begin{equation}
                \mathbf{G}_{b,c}
                =
                \mathcal{P}(\mathbf{K}\odot\mathbf{V})_{b,c}
                =
                \frac{1}{HW}
                \sum_{u=1}^{H}
                \sum_{v=1}^{W}
                (\mathbf{K} \odot \mathbf{V})_{b,c,u,v}
            \end{equation}   
            where $\odot$ denotes the Hadamard product. 
            By pooling the element-wise product of $\mathbf{K}$ and $\mathbf{V}$, the model captures a global summary of feature correlations rather than localising specific token-to-token dependencies.
            
            \paragraph{Aperture Modulation}
            Following the aperture metaphor, a gating function modulates the flow of this global context. The queries $\mathbf{Q}$ are ``excited'' via multiplicative modulation with the transformed global descriptor:
            \begin{equation}
            \mathbf{U} = \mathbf{Q} \odot \sigma(\mathrm{Proj}(\mathbf{G})),
            \end{equation}
            where $\sigma(\cdot)$ represents a hardware-friendly activation (e.g., \textit{Hardsigmoid}). This formulation allows the model to prioritise relevant global features while suppressing noise, thereby dynamically gating spatial information. This mechanism achieves linear complexity $\mathcal{O}(BCHW)$, making it well-suited for resource-constrained deployment.

            \paragraph{ConSmax Activation for DPU Deployment}
            To avoid hardware-intensive softmax normalisation, \textbf{Aper-GATES} adopts a lightweight learnable gating function inspired by ConSmax \cite{ConSmax}. Traditional ConSmax replaces softmax with a parametric transformation:
            \begin{equation}
            \mathrm{ConSmax}(x) = \frac{\exp(x - \beta)}{\gamma},
            \end{equation}
            where $\beta$ and $\gamma$ are learnable head-wise parameters.

            In our implementation, the exponential is replaced by a simplified DPU-compatible variant based on depthwise affine transformation and a piecewise-linear \textit{Hardsigmoid}:
            \begin{equation}
            \mathrm{ConSmax}_{\mathrm{DPU}}(x) = \mathrm{Hardsigmoid}(w \cdot x + b).
            \end{equation}
            The parameters $w$ and $b$ are learned channel-wise through a depthwise $1 \times 1$ convolution.

    \subsection{Training and inference}
    \label{sec:training_raw_model}
        
        TriCCOT is a combination of three independent models that predict sequentially. Therefore,  training is performed sequentially, with each model learning from the previous model's predictions. The RPN is first fitted on the detection training set; its outputs on a separate calibration set are then used to fit the conformal corrections \cite{conformal_gentle_introduction}. The conformal predictor guarantees object integrity while avoiding overly large crops that would degrade inference time. In addition, because the conformal predictor is governed solely by Eq.\ref{eq:conformal} during inference, it is both lightweight and computationally inexpensive to implement. Finally, Aper-GATES is trained from crops generated by the frozen RPN and conformal stages, using the same train/validation split as the RPN. The hyperparameters used to train TriCCOT are available in Tab.~\ref{tab:parameters}. 

        \TabTrainingParams

\subsection{Datasets}

The DIOR dataset \cite{DIOR} is chosen to evaluate the performance of TriCCOT, a widely used Earth-observation benchmark comprising 23,463 optical remote-sensing images across 20 object classes.  
To assess robustness to degraded imagery, we also generate a degraded version of DIOR by applying spatial blur and signal-dependent noise that are representative of raw satellite imaging conditions. The degradation process approximately reverses restoration operations typically applied to satellite imagery \cite{raw_detection_edhpc}.

In addition, we evaluated TriCCOT on real raw satellite imagery with the VENµS raw images for vessel detection (VDVRaw) dataset \cite{venus_dataset_full_author}. VDVRaw contains 282 multispectral images divided into 4\,238 patches affected by various sensor and acquisition artefacts, including striping noise, band misalignment, radiometric noise, blur, and stray light. It contains a total of 3\,827 annotated objects across 17 classes.

\subsubsection{Degradation simulation}

    The degraded DIOR dataset combines spatial and radiometric degradations representative of realistic imaging systems.
    
    \paragraph{Spatial degradation via MTF}
    Spatial resolution loss is modelled using a parametric Modulation Transfer Function (MTF) combining optical blur and sensor sampling:
    \begin{equation}
    \mathrm{MTF}(f_x,f_y)
    =
    e^{-\gamma f_r}\mathrm{sinc}(f_x)\mathrm{sinc}(f_y),
    \qquad
    f_r=\sqrt{f_x^2+f_y^2},
    \end{equation}
    where $\gamma$ is calibrated from a user-defined Nyquist value $MTF_{Nyq}$,
    \begin{equation}
    \gamma=-2\log\left(
    \frac{\mathrm{MTF}_{\mathrm{Nyq}}}{\mathrm{sinc}(0.5)}
    \right).
    \end{equation}
    The corresponding point-spread function is obtained by inverse Fourier transform and convolved independently with each image channel. In our work, $MTF@Nyquist \in [1\%, 2\%]$.
    
    \paragraph{Signal-dependent noise}
    Radiometric degradation is modelled using signal-dependent Gaussian noise,
    \begin{equation}
    \sigma^2(L)=\alpha L+\beta,
    \end{equation}
    where $L$ denotes luminance. The parameters $\alpha$ and $\beta$ are determined from two reference luminance--SNR pairs using
    \begin{equation}
    \sigma_i^2 =
    \left(
    \frac{L_i}{10^{\mathrm{SNR}_{i,\mathrm{dB}}/20}}
    \right)^2
    =\alpha L_i+\beta.
    \end{equation}
    Noise with standard deviation
    \begin{equation}
    \sigma(L)=\sqrt{\max(0,\alpha L+\beta)}
    \end{equation}
    is then sampled independently for each pixel. In our work, the SNR values of the simulated images for two points of luminance $L_0=25 W/m^{2}/sr/\si{\micro\metre} $ and $L_1=100 W/m^{2}/sr/\si{\micro\metre}$ are set between [25dB, 35dB] and [65dB, 75dB], respectively. 
\section{Results}

    \subsection{Ablation study}
    To ensure that all components of TriCCOT contribute to detection, we performed an ablation study on the original DIOR dataset. Aper-GATES is replaced with a traditional ViT implementation, ConSmax is replaced with a plain Hardsigmoid without learnable $w,b$, the multi-resolution patch embedding is replaced with a fixed 128x128 resize, and we completely removed the conformal predictor. The results of the ablation study are available in Tab.~\ref{tab:triccot_ablation}. 
    
    First, we observe that the full implementation of TriCCOT has the best performance on all criteria. This shows that each module contributes to improving detection quality. Removing Aper-GATES results in only a modest reduction in performance, with a decrease of only 0.82 percentage points in mAP@50. This is because Aper-GATES does not aim to improve the attention mechanism, but rather to enable its implementation on accelerated embedded hardware. Moreover, we can see that removing the conformal predictor has a significant impact on performance, with a 7.39-percentage-point decrease in mAP@50. This result shows that the context provided by enlarged bounding boxes helps the transformer process the crop, particularly for misaligned predictions, as demonstrated by the gap in mAP@50.
    
    \TabTriCCOTAblation

    \subsection{Detection performance on the DIOR and VDVRaw datasets}

    TriCCOT is evaluated on both the original DIOR dataset and the raw simulated DIOR dataset. In addition, we evaluated a two-stage detector with Faster R-CNN \cite{faster_RCNN}, lightweight one-stage detectors with YOLOX-S \cite{yolox} and NanoDet-Plus \cite{nanodet}, and a real-time transformer-based detector with RT-DETR \cite{rt_detr}. This comparison covers a broad range of architectures. Tab.~\ref{tab:benchmark_all} reports the results for all models on the three datasets.

    On the original DIOR dataset, RT-DETR achieves the best overall detection performance. Among FPGA-compatible models, YOLOX-S achieves the best mAP@50, with 78.7\%, while NanoDet-Plus achieves the best mAP@50:95, with 47.9\%. Nevertheless, TriCCOT remains close to these results, with a maximum difference of less than 2.5 percentage points for both mAP@50 and mAP@50:95, while requiring only 3.5M parameters and 8.2 GFLOPs. In contrast, DETR achieves 49.9\% mAP@50 and 28.7\% mAP@50:95, highlighting the higher training and deployment costs of a standard transformer detector in this setting, as it must learn spatial relationships \cite{ViT_vs_CNN_biases}.

    To examine the behaviour of TriCCOT beyond aggregate metrics, Tab.~\ref{tab:per_class_dior_small} reports AP@50 for representative DIOR categories spanning different object types and performance levels. Performance ranges from 43.09\% for Bridge to 91.92\% for Airplane, indicating that TriCCOT achieves competitive detections across diverse categories rather than relying on a few high-performing classes.

    Evaluation on the raw DIOR dataset shows that RT-DETR still achieves the best overall results. However, among FPGA-compatible models, TriCCOT obtains the best performance on both metrics, with 74.8\% mAP@50 and 42.2\% mAP@50:95. Notably, the transformer-based architectures (TriCCOT, RT-DETR and DETR) experience the smallest decline in performance when evaluated on the degraded rather than the original dataset,  supporting the hypothesis that attention-based mechanisms improve robustness to image degradations.

    Finally, evaluation on VDVRaw confirms this trend on real raw imagery. Although RT-DETR remains the best-performing model overall, TriCCOT achieves the best results among FPGA-compatible models, reaching 25.3\% mAP@50 and 18.0\% mAP@50:95. These results show that TriCCOT provides a favourable trade-off between robustness to raw-image degradations and embedded deployability.
    
    Furthermore, Fig.~\ref{fig:predictions_and_crops} shows examples of TriCCOT detections for both the original and raw datasets, with examples of the impact of the conformal predictor on the crops. We see that TriCCOT produces qualitatively accurate detections with accurate bounding boxes and class predictions. In these examples, the degradations have little impact on detection. The resulting crops also illustrate the additional context provided by the bounding boxes generated by the conformal predictor.

    \TabPerClassSmall

    \TabBenchmark

    \FigPreds
    
    \subsection{FPGA deployment}
        Finally, TriCCOT is successfully deployed on a Xilinx Versal VCK190 FPGA. The Versal has gained traction in the space community, thanks to its radiation-tolerant specifications \cite{versal_heavy_ion, versal_heavy_ion_radecs}, and state-of-the-art performance \cite{versal_plane_detection_yolo, versal_ship_detection_yolo}. In this work, we use a Deep Learning Processor Unit (DPU) \cite{versal_dpu_for_CNN}, originally designed to accelerate CNN models. Therefore, unlike other implementations of transformers on FPGA boards \cite{heatVit_hardware_efficient_token_pruning, autoVit_fpga_accelerator, fpga_transformer_accelerator, fpga_vit_accelerator_by_mapping_datapath}, we do not alter the FPGA architecture to fit the attention operations. By doing so, we remain flexible and can easily combine multiple architectures, such as CNNs and transformers. Thanks to the novel attention implementation described in Sec.~\ref{sec:vit}, all convolutional and attentional operations are mapped to the DPU, with none offloaded to the CPU.

        Tab.~\ref{tab:fpga_benchmark_full} compiles the results obtained on the Versal. Here, we focus on edge performance metrics, including inference speed and energy consumption. 
        To better understand the overall inference time of TriCCOT, its processing pipeline is broken down into its main components. The RPN is the most time-consuming stage, accounting for 29.1~ms of the total inference time of 35.5~ms. In contrast, the Aper-GATES classification stage runs in only 3.9~ms on the DPU. Despite being executed on the CPU, the conformal prediction and size-selector stages introduce only limited overhead, with execution times of 2.4~ms and 0.1~ms, respectively. For comparison, we also deployed YOLOX-S on the DPU, while DETR inference was performed on the Versal CPU because the DPU does not support its transformer-based architecture. Thanks to its lightweight architecture, TriCCOT is the fastest model on the Versal, achieving 28.2 FPS on 640x640 patches and enabling real-time processing with transformer architectures.

    %\TabFPGABenchmark
    %\TabFPGABenchmarkTriCCOT
    \TabFPGABenchmarkFull
\section{Discussion}
\label{sec:discussions}

\paragraph{Robustness of TriCCOT under degraded imaging conditions against convolutional approaches}
    While TriCCOT does not achieve the highest detection accuracy on the original DIOR dataset, its main objective is not to maximise performance under ideal image conditions, but to preserve detection quality when onboard imagery is degraded by spatial blur and signal-dependent noise. In this setting, TriCCOT achieves the best performance among the evaluated convolutional methods. In addition, transformer architectures exhibited smaller performance drops between the original and degraded datasets. This supports our hypothesis that combining self-attention and global features improves the model's overall robustness to image degradations.

\paragraph{Contribution of the conformal prediction stage}
    The ablation study highlights the importance of the conformal predictor. Extending the crops improves object coverage and adds contextual information. It also helps remove labelling biases in the dataset. 
    The advantage of using conformal prediction is that it removes the need for arbitrary scaling parameters, compensates for model prediction biases, and provides a mathematically defined coverage guarantee.
    
\paragraph{Role of Aper-GATES in embedded transformer deployment}
    A key contribution of TriCCOT is its improved deployability of transformer-based models on FPGA. Aper-GATES's convolutional reformulation of self-attention enables fast inference. In addition, it is possible to combine both convolutional and attention-based architectures within a single DPU. This design greatly simplifies TriCCOT deployment on the Versal VCK190 and enables simultaneous execution of multiple model architectures without reconfiguring the DPU. Although TriCCOT does not match the performance of large foundation transformer models, it offers a practical trade-off between predictive performance and deployability in embedded systems.

\paragraph{TriCCOT classifier}
    Training a transformer is not trivial, as it typically requires large training datasets to mitigate the lack of local biases. Because the transformer classifier in TriCCOT processes only object crops, most of the input consists of task-relevant information. In our experiments, training Aper-GATES was straightforward and yielded fast convergence. All of TriCCOT's parts (RPN, conformal predictor and Aper-GATES) can therefore be trained in only 4-5~h on a single NVIDIA RTX 4090, compared with approximately 30~h for the transformer baselines considered in our experiments. This reduced training cost is enabled by the smaller feature space and compact architecture. By combining convolutional processing for full-image search with transformer-based classification on regions of interest, TriCCOT supports compact and efficient transformer architectures.

\paragraph{Limitations of sequential training}
    Despite these advantages, TriCCOT introduces additional training complexity. The architecture comprises three components: the RPN, the conformal predictor, and the Aper-GATES classifier. Each module is trained after the previous one, which creates dependencies between stages. As a consequence, biases or errors introduced by the RPN may propagate to the conformal predictor and then to the classifier. Although the conformal stage mitigates incomplete or overly tight proposals, the current sequential training strategy does not allow the full pipeline to be optimised end-to-end. Future work should therefore investigate joint or cooperative training strategies or hard-example mining that allow the proposal and classification stages to adapt to one another during training.
    
    Overall, TriCCOT is competitive with state-of-the-art object detection models on original images, and performs best on noisy images against models deployable on FPGAs. It opens the possibility of a lighter transformer architecture with increased deployment capability on edge devices, specifically for onboard remote sensing applications.
    
\section{Conclusion}

In this work, we introduced TriCCOT, a novel tri-part architecture that combines convolution, conformal prediction, and self-attention for onboard object detection under degraded imaging conditions. The method is based on a convolutional region proposal network, a conformal prediction stage, and Aper-GATES, a hardware-friendly attention-based classifier. This design exploits the localisation efficiency of convolutional detectors, the probabilistic coverage guarantees of conformal prediction, and the robustness of attention-based representations, while remaining compatible with embedded FPGA deployment.

Experiments on the DIOR dataset show that TriCCOT achieves competitive detection performance on clean imagery and outperforms the evaluated convolutional and FPGA-compatible baselines under raw imaging degradations. These results demonstrate that combining convolutional proposal generation with attention-based crop classification improves robustness to spatial blur and signal-dependent noise. Aper-GATES makes TriCCOT fully compatible with the FPGA DPU accelerators by replacing standard attention operations with convolutional projections and hardware-friendly gating. This enables CNN-Transformer inference without modifying the accelerator architecture.

Despite these promising results, several limitations remain. TriCCOT is currently trained sequentially, which may propagate errors between the RPN, conformal predictor, and classifier. Future work will investigate joint or cooperative training strategies through hard-example mining \cite{hard_mining_for_single_shot, hard_mining_object_detection} to better optimise the full detection pipeline.

We will also investigate adapting the RPN to rotated bounding-box tasks. Indeed, experiments with remote sensing imagery show that similar objects can exhibit different rotations (such as tennis courts). While the size information is preserved, the ratio can vary greatly. Using rotated bounding boxes could benefit the classifier by harmonising bounding-box ratios across the same classes.

Finally, the proposed region proposal network is based on a CSPDarknet backbone. This sets a hard limit on the total number of parameters and the total size of our model. Ongoing work focuses on replacing CSPDarknet with a lighter convolutional backbone, such as MobileNet \cite{mobilenet}, thereby drastically reducing TriCCOT's overall size and greatly improving inference speed.

\bibliography{main}

@INPROCEEDINGS{versal_spaice_project,
  author={Garcés-Socarrás, Luis M. and Cuiman, Raudel and Ortiz, Flor and Vásquez-Peralvo, Juan A. and González-Rios, Jorge L. and Chehaitly, Mouhamad and Kazanskii, Arkadii and Malmir, Sahar and Nik, Amirhossein and Thoemel, Jan and Kumar, Sumit and Kuhfuss, Marcele and Varadajulu, Swetha and Lagunas, Eva and Duncan, Juan C. M. and Querol, Jorge and Chatzinotas, Symeon},
  booktitle={2025 European Data Handling \& Data Processing Conference (EDHPC)}, 
  title={Onboard Machine Learning for Satellite Edge Computing: The SPAICE Project Use Case}, 
  year={2025},
  volume={},
  number={},
  pages={1-8},

  doi={},
  }

@INPROCEEDINGS{versal_plane_detection_yolo,
  author={Brown, Jacob and Yates, Colton and Goeders, Jeffrey and Wirthlin, Michael},
  booktitle={2025 IEEE Aerospace Conference}, 
  title={RarePlanes Detection Using YOLOv5 on the Versal Adaptive SoC}, 
  year={2025},
  volume={},
  number={},
  pages={1-9},
  doi={10.1109/AERO63441.2025.11068464}}

@INPROCEEDINGS{versal_ship_detection_yolo,
  author={Ibrahim, Younis and Chen, Li and Haonan, Tian},
  booktitle={2022 14th International Conference on Computational Intelligence and Communication Networks (CICN)}, 
  title={Deep Learning-based Ship Detection on FPGAs}, 
  year={2022},
  volume={},
  number={},
  pages={454-459},
  doi={10.1109/CICN56167.2022.10008312}}

@misc{versal_dpu_for_CNN,
      title={DPUV4E: High-Throughput DPU Architecture Design for CNN on Versal ACAP}, 
      author={Guoyu Li and Pengbo Zheng and Jian Weng and Enshan Yang},
      year={2025},
      eprint={2506.11441},
      archivePrefix={arXiv},
      primaryClass={cs.AR},
      url={https://arxiv.org/abs/2506.11441}, 
}

@article{vessel_detection_yolo11_sentinel2,
title = {Mapping recreational marine traffic from Sentinel-2 imagery using YOLO object detection models},
journal = {Remote Sensing of Environment},
volume = {326},
pages = {114791},
year = {2025},
issn = {0034-4257},
doi = {https://doi.org/10.1016/j.rse.2025.114791},
url = {https://www.sciencedirect.com/science/article/pii/S0034425725001956},
author = {Janne Mäyrä and Elina A. Virtanen and Ari-Pekka Jokinen and Joni Koskikala and Sakari Väkevä and Jenni Attila},
}

@ARTICLE{remote_sensing_review,
  author={Zhang, Bing and Wu, Yuanfeng and Zhao, Boya and Chanussot, Jocelyn and Hong, Danfeng and Yao, Jing and Gao, Lianru},
  journal={IEEE Journal of Selected Topics in Applied Earth Observations and Remote Sensing}, 
  title={Progress and Challenges in Intelligent Remote Sensing Satellite Systems}, 
  year={2022},
  volume={15},
  number={},
  pages={1814-1822},
  doi={10.1109/JSTARS.2022.3148139}}

@ARTICLE{cloud_detection_onboard,
  author={Aybar, Cesar and Mateo-García, Gonzalo and Acciarini, Giacomo and Růžička, Vít and Meoni, Gabriele and Longépé, Nicolas and Gómez-Chova, Luis},
  journal={IEEE Journal of Selected Topics in Applied Earth Observations and Remote Sensing}, 
  title={Onboard Cloud Detection and Atmospheric Correction With Efficient Deep Learning Models}, 
  year={2024},
  volume={17},
  number={},
  pages={19518-19529},
  doi={10.1109/JSTARS.2024.3480520}}

@article{opssat_meoni,
  title={The OPS-SAT case: A data-centric competition for onboard satellite image classification},
  author={Meoni, Gabriele and M{\"a}rtens, Marcus and Derksen, Dawa and See, Kenneth and Lightheart, Toby and S{\'e}cher, Anthony and Martin, Arnaud and Rijlaarsdam, David and Fanizza, Vincenzo and Izzo, Dario},
  journal={Astrodynamics},
  volume={8},
  number={4},
  pages={507--528},
  year={2024},
  publisher={Springer}
}

@ARTICLE{irma_imagini,

  author={Goudemant, Thomas and Francesconi, Benjamin and Aubrun, Michelle and Kervennic, Erwann and Grenet, Ingrid and Bobichon, Yves and Bellizzi, Marjorie},
  journal={IEEE Journal of Selected Topics in Applied Earth Observations and Remote Sensing}, 
  title={Onboard Anomaly Detection for Marine Environmental Protection}, 
  year={2024},
  volume={17},
  number={},
  pages={7918-7931},
  doi={10.1109/JSTARS.2024.3382394}}

@inproceedings{CIAR,
  TITLE = {{D{\'e}tection de navires embarquable {\`a} bord de satellites}},
  AUTHOR = {Goudemant, Thomas and Francesconi, Benjamin and Farhat, Houssem and Daniel, Lionel and Thiery, Olivier and Kervennic, Erwann and Girard, Adrien and Mzoughi, Seif},
  URL = {https://hal.science/hal-03881738},
  BOOKTITLE = {{Actes de la 4{\`e}me Conference on Artificial Intelligence for Defense (CAID 2022)}},
  ADDRESS = {Rennes, France},
  ORGANIZATION = {{DGA Ma{\^i}trise de l'Information}},
  SERIES = {Actes de la 4{\`e}me Conference on Artificial Intelligence for Defense (CAID 2022)},
  YEAR = {2022},
  MONTH = Nov,
  HAL_ID = {hal-03881738},
  HAL_VERSION = {v1},
}

@Article{onboard_oil_spill_detection,
AUTHOR = {Diana, Lorenzo and Xu, Jia and Fanucci, Luca},
TITLE = {Oil Spill Identification from SAR Images for Low Power Embedded Systems Using CNN},
JOURNAL = {Remote Sensing},
VOLUME = {13},
YEAR = {2021},
NUMBER = {18},
ARTICLE-NUMBER = {3606},
URL = {https://www.mdpi.com/2072-4292/13/18/3606},
ISSN = {2072-4292},
DOI = {10.3390/rs13183606}
}

@misc{mobilenet,
      title={MobileNets: Efficient Convolutional Neural Networks for Mobile Vision Applications}, 
      author={Andrew G. Howard and Menglong Zhu and Bo Chen and Dmitry Kalenichenko and Weijun Wang and Tobias Weyand and Marco Andreetto and Hartwig Adam},
      year={2017},
      eprint={1704.04861},
      archivePrefix={arXiv},
      primaryClass={cs.CV},
      url={https://arxiv.org/abs/1704.04861}, 
}

@article{DIOR,
   title={Object detection in optical remote sensing images: A survey and a new benchmark},
   volume={159},
   ISSN={0924-2716},
   url={http://dx.doi.org/10.1016/j.isprsjprs.2019.11.023},
   DOI={10.1016/j.isprsjprs.2019.11.023},
   journal={ISPRS Journal of Photogrammetry and Remote Sensing},
   publisher={Elsevier BV},
   author={Li, Ke and Wan, Gang and Cheng, Gong and Meng, Liqiu and Han, Junwei},
   year={2020},
   month=jan, pages={296–307} }

@misc{hard_mining_object_detection,
      title={Training Region-based Object Detectors with Online Hard Example Mining}, 
      author={Abhinav Shrivastava and Abhinav Gupta and Ross Girshick},
      year={2016},
      eprint={1604.03540},
      archivePrefix={arXiv},
      primaryClass={cs.CV},
      url={https://arxiv.org/abs/1604.03540}, 
}

@misc{hard_mining_for_single_shot,
      title={Improved Hard Example Mining Approach for Single Shot Object Detectors}, 
      author={Aybora Koksal and Onder Tuzcuoglu and Kutalmis Gokalp Ince and Yoldas Ataseven and A. Aydin Alatan},
      year={2022},
      eprint={2202.13080},
      archivePrefix={arXiv},
      primaryClass={cs.CV},
      url={https://arxiv.org/abs/2202.13080}, 
}

@misc{CvT_convolutional_vision_transformer,
      title={CvT: Introducing Convolutions to Vision Transformers}, 
      author={Haiping Wu and Bin Xiao and Noel Codella and Mengchen Liu and Xiyang Dai and Lu Yuan and Lei Zhang},
      year={2021},
      eprint={2103.15808},
      archivePrefix={arXiv},
      primaryClass={cs.CV},
      url={https://arxiv.org/abs/2103.15808}, 
}

@misc{CMT_convolutional_NN_meets_transformers,
      title={CMT: Convolutional Neural Networks Meet Vision Transformers}, 
      author={Jianyuan Guo and Kai Han and Han Wu and Yehui Tang and Xinghao Chen and Yunhe Wang and Chang Xu},
      year={2022},
      eprint={2107.06263},
      archivePrefix={arXiv},
      primaryClass={cs.CV},
      url={https://arxiv.org/abs/2107.06263}, 
}

@misc{linear_attention,
      title={Linear attention is (maybe) all you need (to understand transformer optimization)}, 
      author={Kwangjun Ahn and Xiang Cheng and Minhak Song and Chulhee Yun and Ali Jadbabaie and Suvrit Sra},
      year={2024},
      eprint={2310.01082},
      archivePrefix={arXiv},
      primaryClass={cs.LG},
      url={https://arxiv.org/abs/2310.01082}, 
}

@misc{flexivit,
      title={FlexiViT: One Model for All Patch Sizes}, 
      author={Lucas Beyer and Pavel Izmailov and Alexander Kolesnikov and Mathilde Caron and Simon Kornblith and Xiaohua Zhai and Matthias Minderer and Michael Tschannen and Ibrahim Alabdulmohsin and Filip Pavetic},
      year={2023},
      eprint={2212.08013},
      archivePrefix={arXiv},
      primaryClass={cs.CV},
      url={https://arxiv.org/abs/2212.08013}, 
}

@misc{dinov2,
      title={DINOv2: Learning Robust Visual Features without Supervision}, 
      author={Maxime Oquab and Timothée Darcet and Théo Moutakanni and Huy Vo and Marc Szafraniec and Vasil Khalidov and Pierre Fernandez and Daniel Haziza and Francisco Massa and Alaaeldin El-Nouby and Mahmoud Assran and Nicolas Ballas and Wojciech Galuba and Russell Howes and Po-Yao Huang and Shang-Wen Li and Ishan Misra and Michael Rabbat and Vasu Sharma and Gabriel Synnaeve and Hu Xu and Hervé Jegou and Julien Mairal and Patrick Labatut and Armand Joulin and Piotr Bojanowski},
      year={2024},
      eprint={2304.07193},
      archivePrefix={arXiv},
      primaryClass={cs.CV},
      url={https://arxiv.org/abs/2304.07193}, 
}

@misc{swin_transformer,
      title={Swin Transformer: Hierarchical Vision Transformer using Shifted Windows}, 
      author={Ze Liu and Yutong Lin and Yue Cao and Han Hu and Yixuan Wei and Zheng Zhang and Stephen Lin and Baining Guo},
      year={2021},
      eprint={2103.14030},
      archivePrefix={arXiv},
      primaryClass={cs.CV},
      url={https://arxiv.org/abs/2103.14030}, 
}

@misc{MAE_masked_autoencoders,
      title={Masked Autoencoders Are Scalable Vision Learners}, 
      author={Kaiming He and Xinlei Chen and Saining Xie and Yanghao Li and Piotr Dollár and Ross Girshick},
      year={2021},
      eprint={2111.06377},
      archivePrefix={arXiv},
      primaryClass={cs.CV},
      url={https://arxiv.org/abs/2111.06377}, 
}

@misc{DETR,
      title={End-to-End Object Detection with Transformers}, 
      author={Nicolas Carion and Francisco Massa and Gabriel Synnaeve and Nicolas Usunier and Alexander Kirillov and Sergey Zagoruyko},
      year={2020},
      eprint={2005.12872},
      archivePrefix={arXiv},
      primaryClass={cs.CV},
      url={https://arxiv.org/abs/2005.12872}, 
}

@article{ConvVit,
   title={ConViT: improving vision transformers with soft convolutional inductive biases*},
   volume={2022},
   ISSN={1742-5468},
   url={http://dx.doi.org/10.1088/1742-5468/ac9830},
   DOI={10.1088/1742-5468/ac9830},
   number={11},
   journal={Journal of Statistical Mechanics: Theory and Experiment},
   publisher={IOP Publishing},
   author={d’Ascoli, Stéphane and Touvron, Hugo and Leavitt, Matthew L and Morcos, Ari S and Biroli, Giulio and Sagun, Levent},
   year={2022},
}

@misc{ViT_vs_CNN_biases,
      title={An Image is Worth 16x16 Words: Transformers for Image Recognition at Scale}, 
      author={Alexey Dosovitskiy and Lucas Beyer and Alexander Kolesnikov and Dirk Weissenborn and Xiaohua Zhai and Thomas Unterthiner and Mostafa Dehghani and Matthias Minderer and Georg Heigold and Sylvain Gelly and Jakob Uszkoreit and Neil Houlsby},
      year={2021},
      eprint={2010.11929},
      archivePrefix={arXiv},
      primaryClass={cs.CV},
      url={https://arxiv.org/abs/2010.11929}, 
}

@misc{causal_attention_in_VLM,
      title={Causal Attention for Vision-Language Tasks}, 
      author={Xu Yang and Hanwang Zhang and Guojun Qi and Jianfei Cai},
      year={2021},
      eprint={2103.03493},
      archivePrefix={arXiv},
      primaryClass={cs.CV},
      url={https://arxiv.org/abs/2103.03493}, 
}

@misc{attention_all_you_need,
      title={Attention Is All You Need}, 
      author={Ashish Vaswani and Noam Shazeer and Niki Parmar and Jakob Uszkoreit and Llion Jones and Aidan N. Gomez and Lukasz Kaiser and Illia Polosukhin},
      year={2023},
      eprint={1706.03762},
      archivePrefix={arXiv},
      primaryClass={cs.CL},
      url={https://arxiv.org/abs/1706.03762}, 
}

@misc{non_local_nn,
      title={Non-local Neural Networks}, 
      author={Xiaolong Wang and Ross Girshick and Abhinav Gupta and Kaiming He},
      year={2018},
      eprint={1711.07971},
      archivePrefix={arXiv},
      primaryClass={cs.CV},
      url={https://arxiv.org/abs/1711.07971}, 
}

@misc{gate_conv_network,
      title={GCNet: Non-local Networks Meet Squeeze-Excitation Networks and Beyond}, 
      author={Yue Cao and Jiarui Xu and Stephen Lin and Fangyun Wei and Han Hu},
      year={2019},
      eprint={1904.11492},
      archivePrefix={arXiv},
      primaryClass={cs.CV},
      url={https://arxiv.org/abs/1904.11492}, 
}

@INPROCEEDINGS{gated_network,
  author={Hu, Jie and Shen, Li and Sun, Gang},
  booktitle={2018 IEEE/CVF Conference on Computer Vision and Pattern Recognition}, 
  title={Squeeze-and-Excitation Networks}, 
  year={2018},
  volume={},
  number={},
  pages={7132-7141},
  doi={10.1109/CVPR.2018.00745}}

@misc{fpga_vit_accelerator_by_mapping_datapath,
      title={Refining Datapath for Microscaling ViTs}, 
      author={Can Xiao and Jianyi Cheng and Aaron Zhao},
      year={2025},
      eprint={2505.22194},
      archivePrefix={arXiv},
      primaryClass={cs.AR},
      url={https://arxiv.org/abs/2505.22194}, 
}

@article{fpga_transformer_accelerator,
   title={A runtime-adaptive transformer neural network accelerator on FPGAs},
   volume={120},
   ISSN={0141-9331},
   url={http://dx.doi.org/10.1016/j.micpro.2025.105223},
   DOI={10.1016/j.micpro.2025.105223},
   journal={Microprocessors and Microsystems},
   publisher={Elsevier BV},
   author={Kabir, Ehsan and Bakos, Jason D. and Andrews, David and Huang, Miaoqing},
   year={2026},
   month=feb, pages={105223} }

@misc{autoVit_fpga_accelerator,
      title={Auto-ViT-Acc: An FPGA-Aware Automatic Acceleration Framework for Vision Transformer with Mixed-Scheme Quantization}, 
      author={Zhengang Li and Mengshu Sun and Alec Lu and Haoyu Ma and Geng Yuan and Yanyue Xie and Hao Tang and Yanyu Li and Miriam Leeser and Zhangyang Wang and Xue Lin and Zhenman Fang},
      year={2022},
      eprint={2208.05163},
      archivePrefix={arXiv},
      primaryClass={cs.CV},
      url={https://arxiv.org/abs/2208.05163}, 
}

@misc{heatVit_hardware_efficient_token_pruning,
      title={HeatViT: Hardware-Efficient Adaptive Token Pruning for Vision Transformers}, 
      author={Peiyan Dong and Mengshu Sun and Alec Lu and Yanyue Xie and Kenneth Liu and Zhenglun Kong and Xin Meng and Zhengang Li and Xue Lin and Zhenman Fang and Yanzhi Wang},
      year={2023},
      eprint={2211.08110},
      archivePrefix={arXiv},
      primaryClass={cs.AR},
      url={https://arxiv.org/abs/2211.08110}, 
}

@INPROCEEDINGS{raw_detection_edhpc,
  author={Dorise, Adrien and Bellizzi, Marjorie and Girard, Adrien and Francesconi, Benjamin and May, Stéphane},
  booktitle={2025 European Data Handling and Data Processing Conference (EDHPC)}, 
  title={Explaining raw data complexity to improve satellite onboard processing}, 
  year={2025},
  volume={},
  number={},
  pages={1-8},
  doi={}}

@INPROCEEDINGS{pyraws_maritime1,
  author={Del Prete, Roberto and Meoni, Gabriele and Salvoldi, Manuel and Barretta, Domenico and Graziano, Maria Daniela and Longépé, Nicolas and Renga, Alfredo},
  booktitle={IGARSS 2024 - 2024 IEEE International Geoscience and Remote Sensing Symposium}, 
  title={Enhanced Maritime Monitoring Via Onboard Processing Of Raw Multi-Spectral Imagery by Deep Learning}, 
  year={2024},
  volume={},
  number={},
  pages={1713-1717},
  doi={10.1109/IGARSS53475.2024.10641068}}

@misc{pyraws_maritime2,
      title={Enhancing Maritime Situational Awareness through End-to-End Onboard Raw Data Analysis}, 
      author={Roberto Del Prete and Manuel Salvoldi and Domenico Barretta and Nicolas Longépé and Gabriele Meoni and Arnon Karnieli and Maria Daniela Graziano and Alfredo Renga},
      year={2024},
      eprint={2411.03403},
      archivePrefix={arXiv},
      primaryClass={cs.CV},
      url={https://arxiv.org/abs/2411.03403}, 
}

@article{pyraws_thraws,
   title={Unlocking the Use of Raw Multispectral Earth Observation Imagery for Onboard Artificial Intelligence},
   volume={17},
   ISSN={2151-1535},
   url={http://dx.doi.org/10.1109/JSTARS.2024.3418891},
   DOI={10.1109/jstars.2024.3418891},
   journal={IEEE Journal of Selected Topics in Applied Earth Observations and Remote Sensing},
   publisher={Institute of Electrical and Electronics Engineers (IEEE)},
   author={Meoni, Gabriele and Prete, Roberto Del and Serva, Federico and De Beusscher, Alix and Colin, Olivier and Longépé, Nicolas},
   year={2024},
   pages={12521–12537} }

@INPROCEEDINGS{versal_heavy_ion,
  author={Perryman, Noah and Wilson, Christopher and George, Alan},
  booktitle={2023 IEEE Aerospace Conference}, 
  title={Evaluation of Xilinx Versal Architecture for Next-Gen Edge Computing in Space}, 
  year={2023},
  volume={},
  number={},
  pages={1-11},
  doi={10.1109/AERO55745.2023.10115906}}

@INPROCEEDINGS{versal_heavy_ion_radecs,
  author={Dufour, Arnaud and Carron, Jérôme and Pierron, François and Fongral, Matthieu and Dangla, David and Bascoul, Guillaume and Bezerra, Françoise and Mekki, Julien and Malou, Florence and Maillard, Pierre},
  booktitle={2021 21th European Conference on Radiation and Its Effects on Components and Systems (RADECS)}, 
  title={7nm FinFET technology heavy ion SEL evaluation using Xilinx Versal as case study}, 
  year={2021},
  volume={},
  number={},
  pages={1-6},
  doi={10.1109/RADECS53308.2021.9954564}}

@ARTICLE{SPPNet,
  author={He, Kaiming and Zhang, Xiangyu and Ren, Shaoqing and Sun, Jian},
  journal={IEEE Transactions on Pattern Analysis and Machine Intelligence}, 
  title={Spatial Pyramid Pooling in Deep Convolutional Networks for Visual Recognition}, 
  year={2015},
  volume={37},
  number={9},
  pages={1904-1916},
  doi={10.1109/TPAMI.2015.2389824}}

@misc{FPN_feature_pyramid_network,
      title={Feature Pyramid Networks for Object Detection}, 
      author={Tsung-Yi Lin and Piotr Dollár and Ross Girshick and Kaiming He and Bharath Hariharan and Serge Belongie},
      year={2017},
      eprint={1612.03144},
      archivePrefix={arXiv},
      primaryClass={cs.CV},
      url={https://arxiv.org/abs/1612.03144}, 
}

@inproceedings{ConSmax,
author = {Liu, Shiwei and Tao, Guanchen and Zou, Yifei and Chow, Derek and Fan, Zichen and Lei, Kauna and Pan, Bangfei and Sylvester, Dennis and Kielian, Gregory and Saligane, Mehdi},
title = {ConSmax: Hardware-Friendly Alternative Softmax with Learnable Parameters},
year = {2025},
isbn = {9798400710773},
publisher = {Association for Computing Machinery},
address = {New York, NY, USA},
url = {https://doi.org/10.1145/3676536.3676766},
doi = {10.1145/3676536.3676766},
booktitle = {Proceedings of the 43rd IEEE/ACM International Conference on Computer-Aided Design},
articleno = {72},
numpages = {9},
location = {Newark Liberty International Airport Marriott, New York, NY, USA},
series = {ICCAD '24}
}

@misc{yolo,
      title={You Only Look Once: Unified, Real-Time Object Detection}, 
      author={Joseph Redmon and Santosh Divvala and Ross Girshick and Ali Farhadi},
      year={2016},
      eprint={1506.02640},
      archivePrefix={arXiv},
      primaryClass={cs.CV},
      url={https://arxiv.org/abs/1506.02640}, 
}

@misc{yolox,
      title={YOLOX: Exceeding YOLO Series in 2021}, 
      author={Zheng Ge and Songtao Liu and Feng Wang and Zeming Li and Jian Sun},
      year={2021},
      eprint={2107.08430},
      archivePrefix={arXiv},
      primaryClass={cs.CV},
      url={https://arxiv.org/abs/2107.08430}, 
}

@misc{RCNN,
      title={Rich feature hierarchies for accurate object detection and semantic segmentation}, 
      author={Ross Girshick and Jeff Donahue and Trevor Darrell and Jitendra Malik},
      year={2014},
      eprint={1311.2524},
      archivePrefix={arXiv},
      primaryClass={cs.CV},
      url={https://arxiv.org/abs/1311.2524}, 
}

@misc{faster_RCNN,
      title={Faster R-CNN: Towards Real-Time Object Detection with Region Proposal Networks}, 
      author={Shaoqing Ren and Kaiming He and Ross Girshick and Jian Sun},
      year={2016},
      eprint={1506.01497},
      archivePrefix={arXiv},
      primaryClass={cs.CV},
      url={https://arxiv.org/abs/1506.01497}, 
}

@misc{FCOS,
      title={FCOS: Fully Convolutional One-Stage Object Detection}, 
      author={Zhi Tian and Chunhua Shen and Hao Chen and Tong He},
      year={2019},
      eprint={1904.01355},
      archivePrefix={arXiv},
      primaryClass={cs.CV},
      url={https://arxiv.org/abs/1904.01355}, 
}

@misc{AI4SPACE_CONVBEERS,
      title={Rethinking Satellite Image Restoration for Onboard AI: A Lightweight Learning-Based Approach}, 
      author={Adrien Dorise and Marjorie Bellizzi and Omar Hlimi},
      year={2026},
      eprint={2604.12807},
      archivePrefix={arXiv},
      primaryClass={cs.CV},
      url={https://arxiv.org/abs/2604.12807}, 
}

@article{conformal_gentle_introduction,
author = {Angelopoulos, Anastasios N. and Bates, Stephen},
title = {Conformal Prediction: A Gentle Introduction},
year = {2023},
issue_date = {Mar 2023},
publisher = {Now Publishers Inc.},
address = {Hanover, MA, USA},
volume = {16},
number = {4},
issn = {1935-8237},
url = {https://doi.org/10.1561/2200000101},
doi = {10.1561/2200000101},
journal = {Found. Trends Mach. Learn.},
month = mar,
pages = {494–591},
numpages = {114}
}

@InProceedings{conformal_andeol,
  title = 	 {Confident Object Detection via Conformal Prediction
 and Conformal Risk Control: an Application to
 Railway Signaling},
  author =       {Andeol, Leo and Fel, Thomas and de Grancey, Florence and Mossina, Luca},
  booktitle = 	 {Proceedings of the Twelfth Symposium on Conformal
 and Probabilistic Prediction with Applications},
  pages = 	 {36--55},
  year = 	 {2023},
  editor = 	 {Papadopoulos, Harris and Nguyen, Khuong An and Boström, Henrik and Carlsson, Lars},
  volume = 	 {204},
  series = 	 {Proceedings of Machine Learning Research},
  month = 	 {13--15 Sep},
  publisher =    {PMLR},
  url = 	 {https://proceedings.mlr.press/v204/andeol23a.html}
}

@inproceedings{conformal_paper,
  TITLE = {{Object Detection With Probabilistic Guarantees}},
  AUTHOR = {de Grancey, Florence and Adam, Jean-Luc and Alecu, Lucian and Gerchinovitz, S{\'e}bastien and Mamalet, Franck and Vigouroux, David},
  URL = {https://hal.science/hal-03769683},
  NOTE = {This preprint has not undergone peer review or any post-submission improvements or corrections. The Version of Record of this contribution will appear in SAFECOMP 2022, LNCS 13415 proceedings.},
  BOOKTITLE = {{SAFECOMP 2022, LNCS 13415}},
  ADDRESS = {M{\"u}nchen, Germany},
  SERIES = {SAFECOMP 2022, LNCS 13415},
  YEAR = {2022},
  MONTH = Sep,
  HAL_ID = {hal-03769683},
  HAL_VERSION = {v1},
}

@inproceedings{puncc_lib,
  title={PUNCC: a Python Library for Predictive Uncertainty Calibration and Conformalization},
  author={Mendil, Mouhcine and Mossina, Luca and Vigouroux, David},
  booktitle={Conformal and Probabilistic Prediction with Applications},
  pages={582--601},
  year={2023},
  organization={PMLR}
}

@inproceedings{puncc_paper,
  title={Robust Gas Demand Forecasting With Conformal Prediction},
  author={Mendil, Mouhcine and Mossina, Luca and Nabhan, Marc and Pasini, Kevin},
  booktitle={Conformal and Probabilistic Prediction with Applications},
  pages={169--187},
  year={2022},
  organization={PMLR}
}

@ARTICLE{venus_dataset_full_author,
  author={Del Prete, Roberto and Salvoldi, Manuel and Barretta, Domenico and Longépé, Nicolas and Meoni, Gabriele and Karnieli, Arnon and Graziano, Maria Daniela and Renga, Alfredo},
  journal={IEEE Journal of Selected Topics in Applied Earth Observations and Remote Sensing}, 
  title={Enhancing Maritime Situational Awareness Through End-to-End Onboard Raw Data Analysis}, 
  year={2025},
  volume={18},
  number={},
  pages={16997-17018},
  doi={10.1109/JSTARS.2025.3584999}}

@misc{nanodet,
    title={NanoDet-Plus: Super fast and high accuracy lightweight anchor-free object detection model.},
    author={RangiLyu},
    howpublished = {\url{https://github.com/RangiLyu/nanodet}},
    year={2021}
}

@misc{rt_detr,
      title={DETRs Beat YOLOs on Real-time Object Detection},
      author={Yian Zhao and Wenyu Lv and Shangliang Xu and Jinman Wei and Guanzhong Wang and Qingqing Dang and Yi Liu and Jie Chen},
      year={2023},
      eprint={2304.08069},
      archivePrefix={arXiv},
      primaryClass={cs.CV}
}
\end{document}